%% file: paper.tex
\documentclass[nohyperref]{article}

\usepackage{microtype}
\usepackage{graphicx}
\usepackage{subfigure}
\usepackage{booktabs} 

\usepackage{multirow}
\usepackage{rotating}
\usepackage{bm}
\usepackage{wrapfig}
\usepackage{afterpage}
\usepackage{hyperref}

\usepackage[accepted]{icml2022}

\usepackage{amsmath}
\DeclareMathOperator*{\argminA}{argmin}
\DeclareMathOperator{\arcosh}{\mathit{arcosh}}

\usepackage{amssymb}
\usepackage{mathtools}
\usepackage{amsthm}

\usepackage[capitalize,noabbrev]{cleveref}
\crefname{section}{Sec.}{Secs.}
\Crefname{section}{Section}{Sections}
\Crefname{table}{Table}{Tables}
\crefname{table}{Tab.}{Tabs.}
\crefname{appendix}{Appx.}{Appxs.}
\crefname{figure}{Fig.}{Figs.}
\crefname{theorem}{Theorem}{Theorem}

\theoremstyle{plain}
\newtheorem{theorem}{Theorem}[section]

\theoremstyle{definition}
\newtheorem{definition}[theorem]{Definition}

\theoremstyle{remark}

\usepackage[textsize=tiny]{todonotes}

\icmltitlerunning{Plug \& Play Attacks: Towards Robust and Flexible Model Inversion Attacks}

\begin{document}

\twocolumn[
\icmltitle{Plug \& Play Attacks: Towards Robust and Flexible Model Inversion Attacks}



\icmlsetsymbol{equal}{*}

\begin{icmlauthorlist}
\icmlauthor{Lukas Struppek}{tud}
\icmlauthor{Dominik Hintersdorf}{tud}
\icmlauthor{Antonio De Almeida Correia}{tud}
\icmlauthor{Antonia Adler}{bw}
\icmlauthor{Kristian Kersting}{tud,cogn_science,hessianai}

\end{icmlauthorlist}

\icmlaffiliation{tud}{Department of Computer Science, Technical University of Darmstadt, Germany.}
\icmlaffiliation{bw}{Universität der Bundeswehr München, Munich, Germany.}

\icmlaffiliation{cogn_science}{Centre for Cognitive Science, TU Darmstadt, Germany.}
\icmlaffiliation{hessianai}{Hessian Center for AI (hessian.AI), Germany}
\icmlcorrespondingauthor{Lukas Struppek}{lukas.struppek@cs.tu-darmstadt.de}

\icmlkeywords{Model Inversion Attacks, Privacy, Security, Deep Learning}

\vskip 0.3in
]



\printAffiliationsAndNotice{}  

\begin{abstract}
\input{0_abstract}
\end{abstract}

\input{1_Introduction}
\input{2_model_inversion}
\input{3_generative_mia}

\input{4_towards_robust_mia}
\input{5_experiments}
\input{6_discussion}

\vskip 0.1in
{\bfseries Reproducibility Statement.} Our source code is publicly at \href{https://github.com/LukasStruppek/Plug-and-Play-Attacks}{https://github.com/LukasStruppek/Plug-and-Play-Attacks} to reproduce the experiments and facilitate further analysis on MIAs. We state all hyperparameters in the Appendix, and with the configuration files provided with our code. 
\vskip 0.1in

{\bfseries Acknowledgments.} The authors thank Daniel Neider and the anonymous reviewers for fruitful comments and discussions. LS further thanks Kuan-Chieh Jackson Wang for his help with setting up the VMI experiments. This work was supported by the German Ministry of Education and Research (BMBF) within the framework program ``Research for Civil Security'' of the German Federal Government, project KISTRA (reference no. 13N15343).
\bibliography{references}
\bibliographystyle{icml2022}

\cleardoublepage

\appendix

\input{A_proofs}
\input{B_experimental_details}
\input{C_datasets}
\input{D_add_results}

\end{document}

%% file: 0_abstract.tex
Model inversion attacks (MIAs) aim to create synthetic images that reflect the class-wise characteristics from a target classifier's private training data by exploiting the model's learned knowledge. Previous research has developed generative MIAs that use generative adversarial networks (GANs) as image priors tailored to a specific target model. This makes the attacks time- and resource-consuming, inflexible, and susceptible to distributional shifts between datasets. To overcome these drawbacks, we present Plug \& Play Attacks, which relax the dependency between the target model and image prior, and enable the use of a single GAN to attack a wide range of targets, requiring only minor adjustments to the attack. Moreover, we show that powerful MIAs are possible even with publicly available pre-trained GANs and under strong distributional shifts, for which previous approaches fail to produce meaningful results. Our extensive evaluation confirms the improved robustness and flexibility of Plug \& Play Attacks and their ability to create high-quality images revealing sensitive class characteristics.

%% file: 1_Introduction.tex
\section{Introduction}
While deep neural networks keep pushing various benchmarks, the privacy and security of these models still receive little attention, aside from adversarial attacks. Many users may mistakenly assume that knowledge learned from training data is safely encoded in a model's weights, so no privacy risks arise from the model. This assumption is wrong and might lead to serious privacy threats. For example, mobile devices support device unlocking and payment approval by facial recognition. If only the face recognition model without information on the user's identity is then somehow leaked or stolen, an adversary might aim to extract the visual characteristics of the user. If the attack is successful, the targeted individual could be identified, leading to a massive compromise of privacy and security.

This paper focuses on these so-called model inversion attacks (MIAs), which intend to create synthetic images that reflect the characteristics of a specific class from a model's private training data. For face recognition, the target model is trained to classify the identities of a set of people. An adversary without any knowledge about the identities but with access to the trained model then tries to create synthetic facial images that share characteristic features with the training identities, such as gender and hair color, and ideally even allows inferring a person's identity. More intuitively, the adversary can be interpreted as a phantom sketch artist who reconstructs faces from a model's extracted knowledge. In applications such as facial recognition, successful attacks could compromise individual privacy.

For attacking deep neural networks, most MIAs use generative adversarial networks (GANs) \cite{goodfellow_gan} as image priors to generate realistic images. Previous MIAs, which we describe in \cref{sec:related_work}, avoided distributional shifts and relied on GANs trained on the same data distribution as the target model~\cite{secret_revealer, knowledge_mia, variational_mia}, used additional input information such as blurred pictures of a person~\cite{secret_revealer}, and tailored the attack and its image prior to specific target models~\cite{knowledge_mia, variational_mia}, restricting the reuse and flexibility of the attacks. Also, all approaches focused on low-resolution images, which limits the quality of the extracted features, and have yet to show their applicability for higher resolutions. We provide a more formal introduction to MIAs and a theoretical consideration of ideal MIAs and possible degradation factors in \cref{sec:generative_mias}.

The present work aims to overcome these drawbacks. We introduce \textit{Plug \& Play Attacks}, which aim at faster, more robust and flexible MIAs that allow us to create high-resolution images while loosening the dependencies between the image prior and the target models. More precisely, we introduce several novelties to (generative) MIAs in \cref{sec:ppa}. First, we tackle the problem of vanishing gradients during the optimization -- a problem ignored by previous approaches -- by proposing a Poincaré loss function instead of the standard cross-entropy loss. We further integrate random transformations into the attack to avoid overfitting results, support the extraction of robust features and mitigate the risk of generating fooling images. Also, we are the first to show the importance of selecting a subset of meaningful samples from the attack results and propose a novel robustness-based selection process.

Our extensive evaluations in \cref{sec:experiments} demonstrate the high efficacy and robustness of our approach and that it also performs well under distributional shifts between datasets, whereas previous approaches fail to produce meaningful results. Also, we can even utilize publicly available, pre-trained GANs for the attacks, avoiding time- and resource-consuming training. 

With this work, we also want to draw attention to the fact that modern neural networks leak more sensitive information than most users might be aware of and that an adversary might exploit it with manageable effort. We discuss ethical considerations, limitations, and future research in \cref{sec:discussion}.

%% file: 2_model_inversion.tex
\section{Model Inversion in Deep Learning}\label{sec:related_work}
Inversion of neural networks can be realized in three different ways: optimization-based, training-based, or architecture-based. The class of optimization-based approaches, usually known as MIAs, tries to reveal class characteristics by creating synthetic model inputs. \citet{privacy_pharmacogenetics} first introduced MIAs against linear regression models. \citet{mi_confidence_fredriskon} later proposed a gradient descent algorithm to exploit the differentiability of neural networks. However, their approach is limited to shallow networks and grayscale inputs and fails on deeper networks.

To enable MIAs for deeper networks, \citet{secret_revealer} proposed to first train a GAN on public data. The GAN is then used as an image prior to restrict the optimization space of generated images. The attack optimizes the GAN's latent input vectors to minimize a cross-entropy loss of the target model's prediction on the generated images. The authors also added a discriminator loss to penalize unrealistic images and took auxiliary knowledge such as blurred images of the target class into account.

\citet{knowledge_mia} built upon this approach and improved the GAN's training process by including soft-labels produced by the target model. To recover the distribution for a target class rather than a single data point, the authors proposed to learn the mean and standard deviation of the latent distribution for each target class modeled by the generator.

Recently, \citet{variational_mia} introduced variational MIAs and formulated a variational objective to account for both diversity and accuracy. The authors trained deep flow models to approximate a separate distribution for each input latent vector in a StyleGAN2~\cite{Karras2019stylegan2} model.

In contrast, training-based approaches interpret the target model as an encoder and train a corresponding decoder network to reconstruct inputs from a model's outputs. It should be noted that not all approaches in this group are intended as privacy attacks. Previous works trained convolutional networks~\cite{inverting_visual_representations, inverting_background_knowledge} and autoregressive neural density models~\cite{inverting_autoregressive} as decoders. A recent work~\cite{inverting_explanations} shows that model explanations improve the inversion results and might harm privacy.

The group of architecture-based model inversion comprises various invertible network architectures that are bijective function approximators and allow the inversion of forward passes. Examples of invertible neural networks include invertible residual networks~\cite{i_resnets}, masked convolutions~\cite{mintnet}, and normalizing flow-based models~\cite{glow}.

In this work, we focus on the group of optimization-based inversion using GANs and will explain the foundations of generative MIAs more formally in the next section.

%% file: 3_generative_mia.tex
\section{Generative Model Inversion Attacks}\label{sec:generative_mias}
We define a target image classification model $M_\mathit{target} \colon X_\mathit{target} \to Y_\mathit{target}$ to be a neural network mapping input images $x \in X_\mathit{target}$ to score vectors $y \in Y_\mathit{target}$. The vector element $M_\mathit{target}(x)_c=y_c \in [0, 1]$ describes the model's prediction score for class $c \in C$. The prediction scores are usually computed by applying a softmax function on the model's output logits $o \in \mathbb{R}^{|C|}$.

In the standard MIA setting, an adversary has full white-box access to $M_\mathit{target}$ with the ability to run an unlimited number of queries and compute gradients but only limited to no information on the learned classes $C$. This is a reasonable assumption, given that hacks, leaks, or data breaches are no rarity in today's world. Also, malicious cloud service providers could gain direct access to the model.

The adversary then tries to create synthetic images $\hat{x}$ that are characteristic for a target class $c$ and potentially leak sensitive information. A naive approach would be to find any image $x^*$ that maximizes $y_c$ by defining an adequate loss function $\mathcal{L}$, e.g., a cross-entropy loss, and then solving
\begin{equation} \label{eq:model_inv_naive}
\begin{split}
x^* = \argminA_{\hat{x}} & \quad \mathcal{L}(M_\mathit{target}, \hat{x}, c).
\end{split}
\end{equation}
While directly optimizing a synthetic image $\hat{x}$ might produce some meaningful results on shallow neural networks~\cite{mi_confidence_fredriskon}, it completely fails on modern, deep neural networks. To overcome this problem, an image prior that captures image statistics from a data distribution $P(X_\mathit{prior})$ could be applied~\cite{secret_revealer}. One option is to use generative adversarial networks (GANs)~\cite{goodfellow_gan}, which consist of a generator $G\colon Z \to X_\mathit{prior}$ and a discriminator $D\colon X_\mathit{prior} \to \mathbb{R}$ network. While $G$ learns to map latent vectors $z \in Z$ to the image space, $D$ tries to distinguish between real samples $x\sim P(X_\mathit{prior})$ and generated samples $G(z)$. \cref{eq:model_inv_naive} can now be extended by $G$ and $D$, and the optimization space gets limited to $Z$:
\begin{equation} \label{eq:gen_model_inv}
\begin{split}
z^* = \argminA_{\hat{z}} & \quad \mathcal{L}(M_\mathit{target}, G, D, \hat{z}, c).
\end{split}
\end{equation}
By solving \cref{eq:gen_model_inv} and optimizing latent vector $\hat{z}$, we might end up with images $x^*=G(z^*)$ for which $M_\mathit{target}$ predicts high scores $y_c$, while $G$ and $D$ assure that ${x^*\sim P(X_\mathit{prior})}$. However, it might still not lead to meaningful results.

\textbf{Ideal MIAs:} To better understand the factors influencing the success of MIAs, we next describe their goals in more detail. For an image distribution $P(X)$, we define $\mathcal{F}=\mathcal{F}(X)$ to be the distribution of human-recognizable features a sample from $X$ can have. We further denote $\mathcal{F}_c = \mathcal{F}(X|c)$ to be the characteristic features for class $c$. For facial images, $\mathcal{F}$ might contain features such as hair color, wrinkles, and interpupillary distance, whereas $P(X)$ also incorporates features not related to a person's identity, such as image background or clothing. This differentiation is important for the distributional shift setting.

We assume characteristic features of two classes $c \neq \tilde{c}$ to be non-identical: $\mathcal{F}_c \neq \mathcal{F}_{\tilde{c}}$. Hence, samples of class $c$ can be characterized only by its feature distribution $\mathcal{F}_c$. Note that feature overlappings between different $\mathcal{F}_c$ are possible. 

A discriminative model $M$ trained on a labeled dataset $(X, C)$ can learn to extract features $\mathcal{F}_M(x)$ from samples $x\in X$ and differentiate between classes $C$ by estimating $M(x)=P(C|\mathcal{F}_M(x))$.

Furthermore, given another sufficiently large dataset $\tilde{X}$, a generative model $G$ can learn to approximate $P(\tilde{X})$ and hence $\mathcal{F}(\tilde{X})$. The model can then generate samples $\hat{x} \sim P(\tilde{X})$ with $\mathcal{F}(\hat{x}) \sim \mathcal{F}(\tilde{X})$. This brings us to the definition of an ideal MIA:

\begin{definition}[Ideal MIA]\label{theo:ideal_mia}
Be ${M\colon X \to C}$ an ideal classifier trained on $(X, C)$, and $G\colon Z \to \tilde{X}$ an ideal generative model trained on $\tilde{X}$ with $\mathcal{F}(\tilde{X}) \approx \mathcal{F}(X)$. By solving $z^*= \argminA_{\hat{z}} \mathcal{L}(M, G, \hat{z}, c)$ with an adequate loss function $\mathcal{L}$ to maximize the prediction score for class $c$, the generated samples $x^*=G(z^*)$ have features $\mathcal{F}(x^*) \approx \mathcal{F}_c$ and, consequently, reveal the characteristics of class c.
\end{definition}

Note that $P(\tilde{X}) \approx P(X)$ does not need to be fulfilled since we are mainly interested in recovering the characteristic features $\mathcal{F}_c$. In our and many previous works, the adversary does not aim to recreate the original training data but to create samples that follow their distribution and reveal characteristics of the targeted identity. A person's identity could be inferred even if the style of synthetic and training samples differ. 

Moreover, in distributional shift settings, such as those studied in this work, there are likely no latent vectors that allow the GAN to generate samples identical or even close to the training data. For example, the StyleGAN2 trained on FFHQ we used for our experiments also generates image backgrounds, whereas samples from the target datasets, such as FaceScrub and CelebA, do not contain any background information. However, MIAs, in general, could also be interpreted as aiming at recovering $P(X|c)$ or specific instances from the training data.

\begin{figure}
\centering
\includegraphics[width=\linewidth]{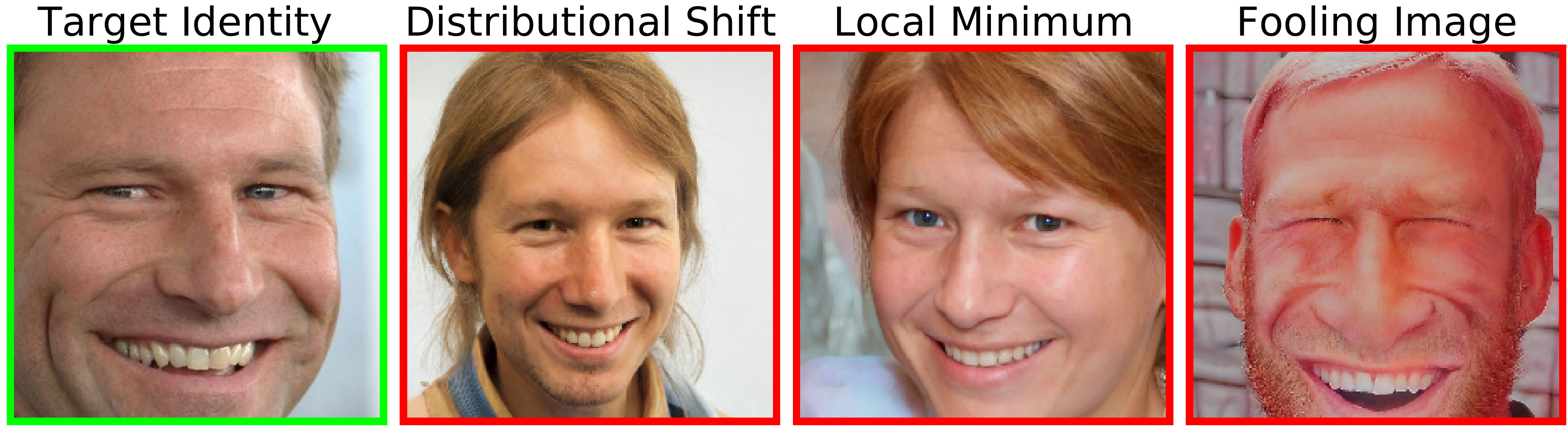}
\vskip -0.1in
\caption{Examples for MIA degradation. Even if none of the three generated images shares the same characteristic features with the targeted identity (left image), all images are classified as the same class. Without further knowledge, an attacker cannot tell such ill-fated attack results apart from meaningful results.
}
\label{fig:mia_problems}
\vskip -0.1in
\end{figure}

\textbf{Degradation Factors for MIAs:} Based on Definition~\ref{theo:ideal_mia}, we outline major degradation causes of MIAs under realistic conditions. First, neither $G$ nor $M_\mathit{target}$ are ideal and decrease the power of MIAs. To distinguish a specific class~$c$ from other samples, $M_\mathit{target}$ only needs to learn feature combinations that are not shared with any other class and might not learn the remaining characteristics of class $c$.

Second, distributional shifts between the datasets complicate MIAs. Given that $P(X_\mathit{target}, C_\mathit{target})$ is the dataset to train $M_\mathit{target}$ and $P(X_\mathit{prior}, C_\mathit{prior})$ the dataset used to train the image prior, previous work assumed that $P(X_\mathit{target}) \approx P(X_\mathit{prior})$ and focused only on label shifts $P(C_\mathit{target}) \neq P(C_\mathit{prior})$. However, our attacks also take covariate shifts $P(X_\mathit{target}) \neq P(X_\mathit{prior})$ into account, which makes the attacks more difficult since the styles of samples from both distributions differ and distract $M_\mathit{target}$ in its predictions on generated samples.

\begin{figure*}[ht]
\centering
\includegraphics[width=\linewidth]{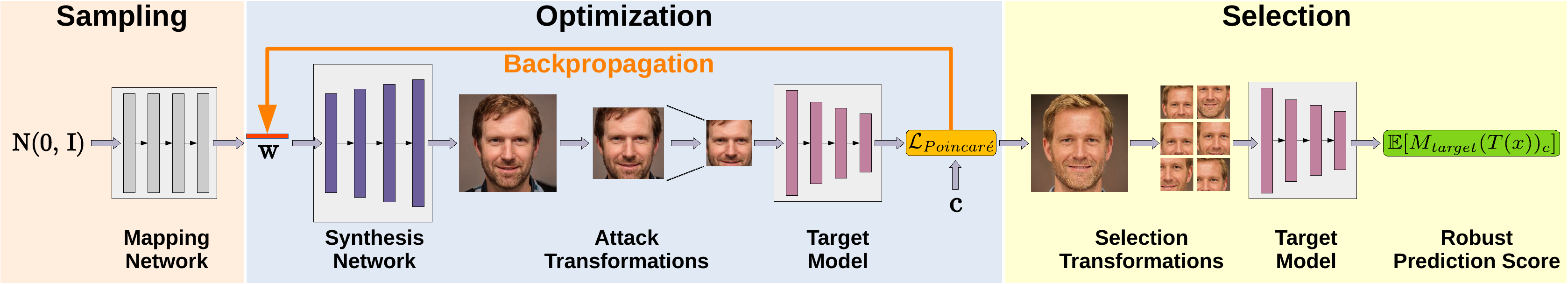}
\vskip -0.1in
\caption{Overview of the three stages of our Plug \& Play Attacks. First, latent vectors are sampled and mapped to their intermediate representation $w$. Images are then generated based on $w$, transformed, and fed into the target model. Finally, a Poincaré loss is computed on the target model's output and target class $c$, and $w$ is updated by back-propagating the loss and performing a gradient descent step. After the optimization is finished, a subset of results is selected based on their robustness against random transformations.}
\label{fig:attack}
\end{figure*}

Third, the MIA's optimization problem is non-convex, and minimization using a gradient descent approach easily ends up in poor local minima. Escaping such minima needs sufficiently large gradients. If the optimization process gets stuck, the generation and extraction of some characteristic features might not be possible. As we show later, a cross-entropy loss might not be the best loss function due to vanishing gradients. We overcome this problem by replacing it with the Poincaré distance.

And last, as various research has pointed out, neural networks are usually not robust and overconfident in their predictions, e.g., on fooling images~\cite{NguyenYC15}, out-of-distribution data~\cite{hein2019, HendrycksG17} or adversarial examples~\cite{szegedy2014intriguing}. Even using an image prior does not prevent the generation of such misleading samples in an MIA. This fact has mostly been ignored by previous work. We propose a simple yet effective selection approach to avoid poorly generated samples.

We visualize poor attack outcomes for distributional shifts, local minima, and fooling images in \cref{fig:mia_problems}. The target model's prediction scores for all three samples are close to 1.0, although the images strongly differ in their characteristic features. While fooling or unrealistic images might be detected by the adversary, it is barely possible to distinguish ill-generated results from images revealing sensitive features. We emphasize that the listed degradation factors are not separate effects but mutually influence each other. See \cref{app:mia_problems_details} for details on the creation of \cref{fig:mia_problems}.

%% file: 4_towards_robust_mia.tex
\section{Towards Robust and Flexible MIAs}\label{sec:ppa}
We now present our main contributions and explain the various components of our Plug \& Play Attacks to make MIAs more robust and applicable in distributional shift settings. See \cref{fig:attack} for an overview of the attack pipeline.

\subsection{Target-Independent Image Priors}\label{sec:image_priors}
We mainly rely on pre-trained StyleGAN2\footnote{During the work on this paper, StyleGAN3~\cite{stylegan3} was published. While we continued to use StyleGAN2, we note that our attack is also compatible with StyleGAN3.}~\cite{Karras2019stylegan2} as image priors to demonstrate that our attack does neither need image priors trained and adapted directly to a specific target model nor any auxiliary inputs -- a generator trained on samples from the same domain, such as facial images for attacking face recognition systems, is sufficient.

The StyleGAN2 generator $G$ consists of two components, a mapping network $G_\mathit{mapping}\colon Z \to W$, which maps random latent vectors $z \in Z$ with $z \sim \mathcal{N}(0, 1)$ into an intermediate latent representation $w \in W$, and a synthesis network ${G_\mathit{synthesis}\colon W \to X_\mathit{prior}}$, which generates images from $w$. The StyleGAN architecture~\cite{Karras_stylegan1} promises a reduced feature entanglement in $W$, which is also beneficial for optimization and facilitates the generation of specific features without affecting other features. For our attacks, we first sample a set of $z \sim \mathcal{N}(0, 1)$, map them with $G_\mathit{mapping}$ to space $W$, and then iteratively modify each $w$ towards our optimization goal.

Unlike previous works, we do not include any discriminator $D$ in our attacks since $D$ would force the generated images to be close to $P(X_\mathit{prior})$ and prevent the attack from approaching $P(X_\mathit{target})$. Moreover, the optimization is guided by a single loss function, which we introduce in \cref{sec:poincare_loss}.

Since $G$ is trained independently of $M_\mathit{target}$ and does not rely on any auxiliary knowledge to generate samples, we can flexibly exchange both $G$ and $M_\mathit{target}$ in our attack pipeline. In particular, it is not necessary to train a separate image prior for each individual target model, but we can re-use each image prior to attack various models trained on different datasets from the same domain.

\subsection{Increasing Robustness by Transformations}\label{sec:augmentation}
As described before, we focus on MIAs with distributional shifts between $P(X_\mathit{target})$ and $P(X_\mathit{prior})$. We mitigate the differences by applying standard image transformations and exploit the fact that many transformations are differentiable and allow gradient computation. We define a single image transformation $t(x)\colon \mathbb{R}^{H\times W \times C} \to \mathbb{R}^{H'\times W' \times C}$, which might include some randomness. Be further $T(x)=t_1(x)\circ \ldots \circ t_m(x)$ a sequential application of a tuple of transformations. During each optimization step, we first generate images, apply the transformations and feed them into the target model to compute its prediction scores. Combined, we compute $M_\mathit{target}(T(G_\mathit{synthesis}(w)))$ during each forward pass.

To adapt the generated images to $P(X_\mathit{target})$, we may apply standard image transformations, such as cropping, scaling, padding, or linear transformations of the pixel values. We expect the attacks to be more successful the closer the transformed images approximate the target distribution.

We further assume the adversary knows the rough style and size of samples from $P(X_\mathit{target})$. However, even without detailed knowledge, by performing the attack with varying parameters and comparing the generated examples, an adversary might still find suitable parameters.

To further reduce the risk that our attack generates misleading images, we include random transformations, such as random cropping or flipping. Our intuition behind this step is that by applying random transformations, our attack not only optimizes a single image but a set of representations based on the same latent vector. We expect the attack not only to extract more robust features but also the results to be less likely out-of-distribution or adversarial examples if a model shows similarly high prediction scores for various transformed versions of an image. In our experiments, we used center cropping and resizing to adjust the generated images, followed by a random cropping step with resizing. Related work utilized image transformations to create robust adversarial examples~\cite{Athalye18} or to visualize learned features in a neural network~\cite{olah2017feature}.

\subsection{Overcoming Vanishing Gradients}\label{sec:poincare_loss}
Previous MIAs relied on a cross-entropy loss $\mathcal{L}_{CE}$ to guide the optimization towards the target class $c$. The derivative of $\mathcal{L}_{CE}$ with respect to the output logit $o_c$ is
\begin{equation}
\begin{aligned}
    \frac{\partial \mathcal{L}_{CE}}{\partial o_c} & = y_c - t_c.
\end{aligned}
\end{equation}
Here, $y_c$ denotes the prediction score for class $c$ and $t_c$ the entry for class $c$ in the one-hot encoded target vector.

\begin{figure}
\centering
\includegraphics[width=\linewidth]{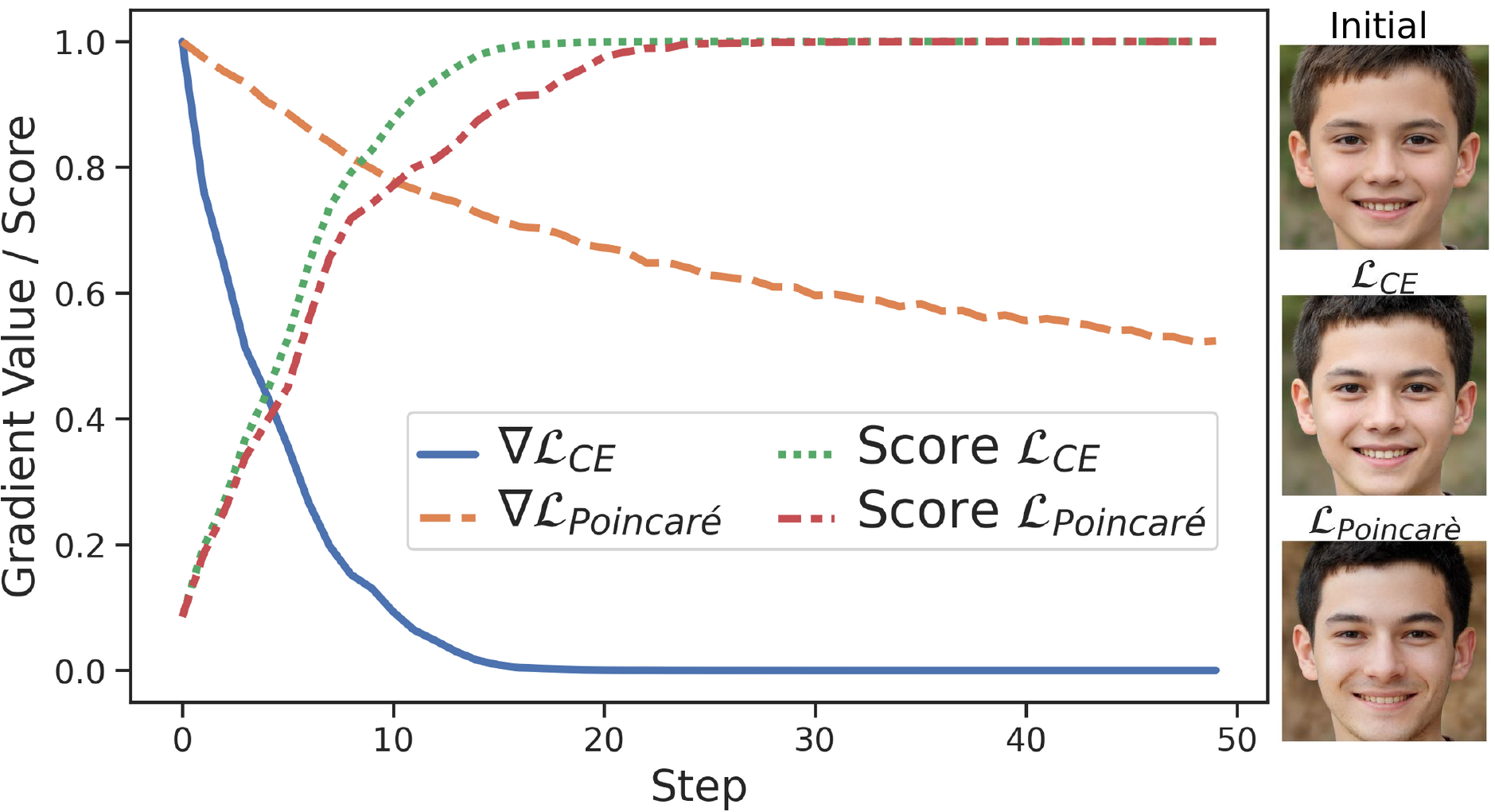}
\vskip -0.1in
\caption{Comparison between $\mathcal{L}_{CE}$ and $\mathcal{L}_{Poincar\acute{e}}$ in terms of average prediction scores and normalized gradient values. The images on the right show that the optimization with $\mathcal{L}_{Poincar\acute{e}}$ escapes the poor local minima, whereas $\mathcal{L}_{CE}$ sticks with the initial features. Still, both optimization processes lead to high prediction scores. The top image depicts the initial sample to optimize.}
\label{fig:loss_comparison}
\vskip -0.1in
\end{figure}

One major drawback of $\mathcal{L}_{CE}$ in MIAs is that the gradient decreases monotonously while the score $y_c$ approaches $t_c$. We prove this statement in~\cref{app:ce_derivative}. It makes optimizing the latent vectors difficult since visual changes to the generated images mainly occur while the target scores are low, and later with increasing scores, the gradients tend to vanish. Early features defined in the latent vectors, and determined by the random sampling process, have a strong impact on the final images and might be adjusted insufficiently to the actual features of the target class during the optimization. It leads to attack outcomes with little meaning for ill-sampled latent vectors and results in poor local minima.

To overcome this problem, we move the optimization to hyperbolic spaces, i.e., non-euclidean spaces with constant negative curvature, and use the Poincaré distance, which origins from hyperbolic geometry, to guide the attack instead. Our approach relies on the Poincaré ball model of hyperbolic geometry and uses its property that the surface area grows exponentially with respect to its radius. Also, the Poincaré space is smooth and differentiable. Related work used hyperbolic geometry to learn hierarchical representations in word embeddings~\cite{poincare, poincare_glove}, optimize variational auto-encoders~\cite{poincare_vae} or create adversarial examples~\cite{transferable_adv_examples}.

We define the Poincaré loss function as the Poincaré distance between two vectors $u, v \in \mathbb{R}^n$ with $\lVert \cdot \rVert_2$ being the Euclidean norm and $\lVert u \rVert_2 < 1$, $\lVert v \rVert_2 < 1$ as
\begin{equation}
\begin{aligned}
    \mathcal{L}_\mathit{Poincar\acute{e}} & = d(u, v) \\
    & =\arcosh\left( 1 + \frac{2 \lVert u-v \rVert_2^2}{(1-\lVert u \rVert_2^2)(1-\lVert v \rVert_2^2)} \right).
\end{aligned}
\end{equation}
We follow \citet{transferable_adv_examples} and set $u$ to be the normalized output logits $u=\frac{o}{\lVert o \rVert_1}$ with $\lVert \cdot \rVert_1$ being the absolute-value norm and $v$ as the one-hot encoded target vector $t$, where we replaced the $1$ by $0.9999$. For vectors with a norm close to one, the distance increases rapidly since the denominator approaches zero. See \cref{app:poincare_derivative} for the mathematical deduction of the derivative.

We compare both loss functions in \cref{fig:loss_comparison} by performing our attack against ten different target classes, with each loss function once. We measured the average prediction scores for the target classes and the gradient absolute-value norms for the generated images. Gradients for $\mathcal{L}_\mathit{Poincar\acute{e}}$ are still present for higher scores, whereas they quickly start to vanish for $\mathcal{L}_\mathit{CE}$. Additionally, we depict a poorly selected starting image, together with the attack results, demonstrating that optimization with $\mathcal{L}_{Poincar\acute{e}}$ induces significant changes to the generated features, whereas $\mathcal{L}_{CE}$ mainly stays close to the original ones. For experimental details on the comparison and further insights, see \cref{app:loss_comparison_details}.

\subsection{Selecting Meaningful Attack Results}\label{sec:sample_selection_process}
\begin{figure}[t]
\centering
\includegraphics[width=0.75\linewidth]{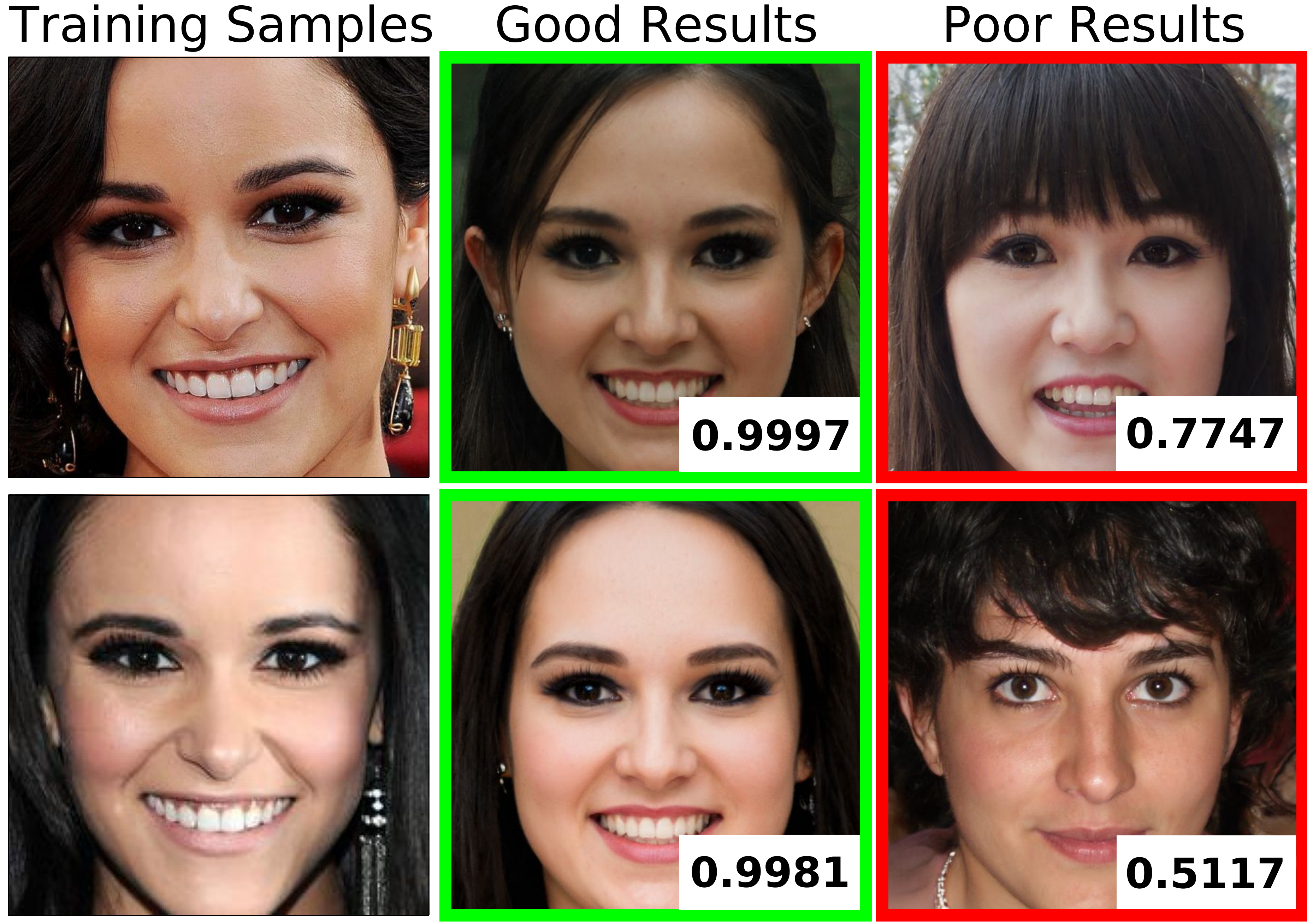}
\vskip -0.1in
\caption{Examples of our sample selection process with their mean prediction scores after transformations were applied. While the mean prediction score for the good attack results (green frame) stays close to $1.0$, the values for the poor results (red frame) drop noticeably, making them easy to separate.}
\label{fig:selection_process}
\end{figure}

Previous work did not pay attention to the selection of initial and optimized latent vectors, and ignored that attack outcomes might be misleading despite regularization. However, we demonstrate that the selection of initial latent vectors to optimize and the final subset of attack results have a strong impact on the effectiveness of MIAs and should not be ignored. At the beginning of the attack, we sample a large number of $z \sim \mathcal{N}(0, 1)$, usually 2,000 vectors, and map them to the intermediate latent space $W$. We then generate images $G_\mathit{synthesis}(w)$ for all $w\in W$, crop and resize them, and compute the mean prediction scores of the images and their horizontally flipped counterpart with $M_{target}$. For each class, we select a subset with a predefined number of vectors $w$ whose corresponding images achieve the highest initial prediction scores. In our experiments, we selected 200 latent vectors to be optimized for each class.

As mentioned before, $M_{target}$ assigns high prediction scores to nearly all attack results, but the images might still not contain any meaningful content in terms of the MIA goals. Thus, we propose a simple yet effective selection process to choose a subset of final images. For the final selection of attack outcomes, our underlying assumption is as follows: Given are two images $x, \tilde{x}$ with high prediction scores $M_\mathit{target}(x)_c \approx M_\mathit{target}(\tilde{x})_c \approx 1.0$ for an arbitrary target class $c$. Be $x$ further a sample representing the characteristics of the target class and $\tilde{x}$ a misleading sample. We assume $\mathbb{E}[M_\mathit{target}(T(x))_c] > \mathbb{E}[M_\mathit{target}(T(\tilde{x}))_c]$ with sufficiently strong image transformations $T$ and, consequently, the prediction scores to be more stable for images representing the correct characteristics. This assumption is inspired by label-only membership inference attacks~\cite{label_only_mem_inf_choquette, label_only_mem_inf_li}, which show that training examples exhibit higher robustness against transformations. We approximate the expected robust prediction scores with a Monte Carlo approach
\newpage

\begin{equation}
    \mathbb{E}[M_\mathit{target}(T(x))_c] \approx \frac{1}{N} \sum_{i=1}^N M_\mathit{target}(T(x))_c
\end{equation}
and apply the random transformations $N=100$ times on the candidate images. We then compute the average prediction scores for the target class and select the $50$ samples with the highest average prediction scores as final attack results. Generally, the transformations used for selection should be different or stronger than the transformation used in the optimization process. Otherwise, poorly generated samples may overfit the target model despite transformations applied. In our experiments, we cut our random patches from the images with random aspect ratios and resized them to match the target model's preferred input size.

\cref{fig:selection_process} visualizes the results of our transformation-based selection process on four attack results targeting the same class. The results visually close to the target identity produce robust prediction scores on $M_\mathit{target}$, while the prediction scores for the ill-generated images drop markedly. Without transformations applied, $M_\mathit{target}$ assigns all four images with a maximum score of $1.0$. Moreover, for the samples from \cref{fig:mia_problems}, the average target scores under random transformations drop to nearly $0.0$, whereas the scores without transformations applied are close to $1.0$.

We also studied other selection approaches based on the robustness against random noise added to the images or their corresponding latent vectors. Also, we investigated robustness against adversarial perturbations~\cite{fgsm}. However, none of these approaches improved the selection results over the proposed transformation-based selection and increased time and resource consumption.

Combining the different parts of our proposed Plug \& Play Attacks, we want to solve the following problem:
\begin{equation} \label{eq:adversary1_problem}
\begin{split}
\min_{\hat{w}} & \quad \mathcal{L}_\mathit{Poincar\acute{e}}(M_\mathit{target}(T(G_{\mathit{synthesis}}(\hat{w}))), c) \\
\textrm{s.t.} & \quad G_{\mathit{synthesis}}(\hat{w}) \in[-1, 1]^{H \times W \times C}.
\end{split}
\end{equation}

%% file: 5_experiments.tex
\section{Experiments}\label{sec:experiments}
We now empirically evaluate the effectiveness of Plug \& Play Attacks and compare it to previous MIA approaches.

\textbf{Experimental Setup.} See \cref{app:experimental_details} for additional details on the experiments. We trained various ResNet~\cite{resnet_he}, ResNeSt~\cite{zhang2020resnest}, and DenseNet~\cite{densenet} models as targets and Inception-v3~\cite{inception_v3} models for evaluation. We trained these models on FaceScrub~\cite{facescrub} and CelebA~\cite{celeba} for facial image classification and Stanford Dogs~\cite{stanford_dogs} for dog breed classification.

We further used publicly available StyleGAN2 models pre-trained on Flickr-Faces-HQ (FFHQ)~\cite{Karras_stylegan1}, MetFaces~\cite{Karras2020ada}, and Animal Faces-HQ Dogs (AFHQ Dogs)~\cite{stargan} as image priors. The faces in FFHQ on one side and CelebA and FaceScrub on the other differ visually significantly, with FFHQ showing a person's full head together with image background at high resolution, and FaceScrub and CelebA mainly containing faces at lower resolutions. Therefore, the attack has to come up with reasonable solutions for missing parts of the head, clothing, and image background. Furthermore, a large distributional shift exists for attacks using the MetFaces prior, a dataset of faces extracted from art. To extend our analyses beyond facial recognition, we also perform attacks against dog breed classifiers. AFHQ Dogs and Stanford Dogs have overlapping dog breeds, but AFHQ contains frontal shots of dogs, whereas Stanford Dogs depicts entire scenarios. Samples and more details are stated in \cref{app:dataset}. 

\textbf{Evaluation Metrics.}\label{sec:metrics}
In line with previous MIA research, we computed various evaluation metrics. First, we trained independent Inception-v3 evaluation models on the target models' training data. We then used the evaluation models to predict the labels on the attack results and computed the top-1 and top-5 accuracy for the targeted classes.

Second, we computed for each generated image the shortest feature distance to any training sample from the target class and stated the average distance $\delta_{eval}$. Distances are measured by the squared $\ell_2$ distance between the activations in the evaluation models' penultimate layers. For facial images, we also used a pre-trained FaceNet~\cite{facenet} to measure the feature distance $\delta_{face}$. Lower values indicate attack results visually closer to training data.

The third metric is the Fréchet inception distance (FID)~\cite{fid_score}, usually used to assess GANs. The FID computes the distance between the feature vectors of images from the target's training data and generated attack results. Feature vectors are extracted by an Inception-v3 model trained on ImageNet~\cite{deng2009imagenet}. A lower FID score indicates a higher similarity between both datasets. See \cref{app:fid_score} for more details, and \cref{app:dataset} for comparison values for the FID score and the feature distances computed on the datasets as baselines.

We further followed Wang et al.~\cite{variational_mia} and computed the improved precision and recall for GANs~\cite{improved_precision}, together with the density and coverage~\cite{coverage_density} on a per-class basis, to evaluate the sample diversity. Our results for these four metrics are stated in \cref{app:add_metric_results}.

\begin{figure}
  \centering
  \includegraphics[width=\linewidth]{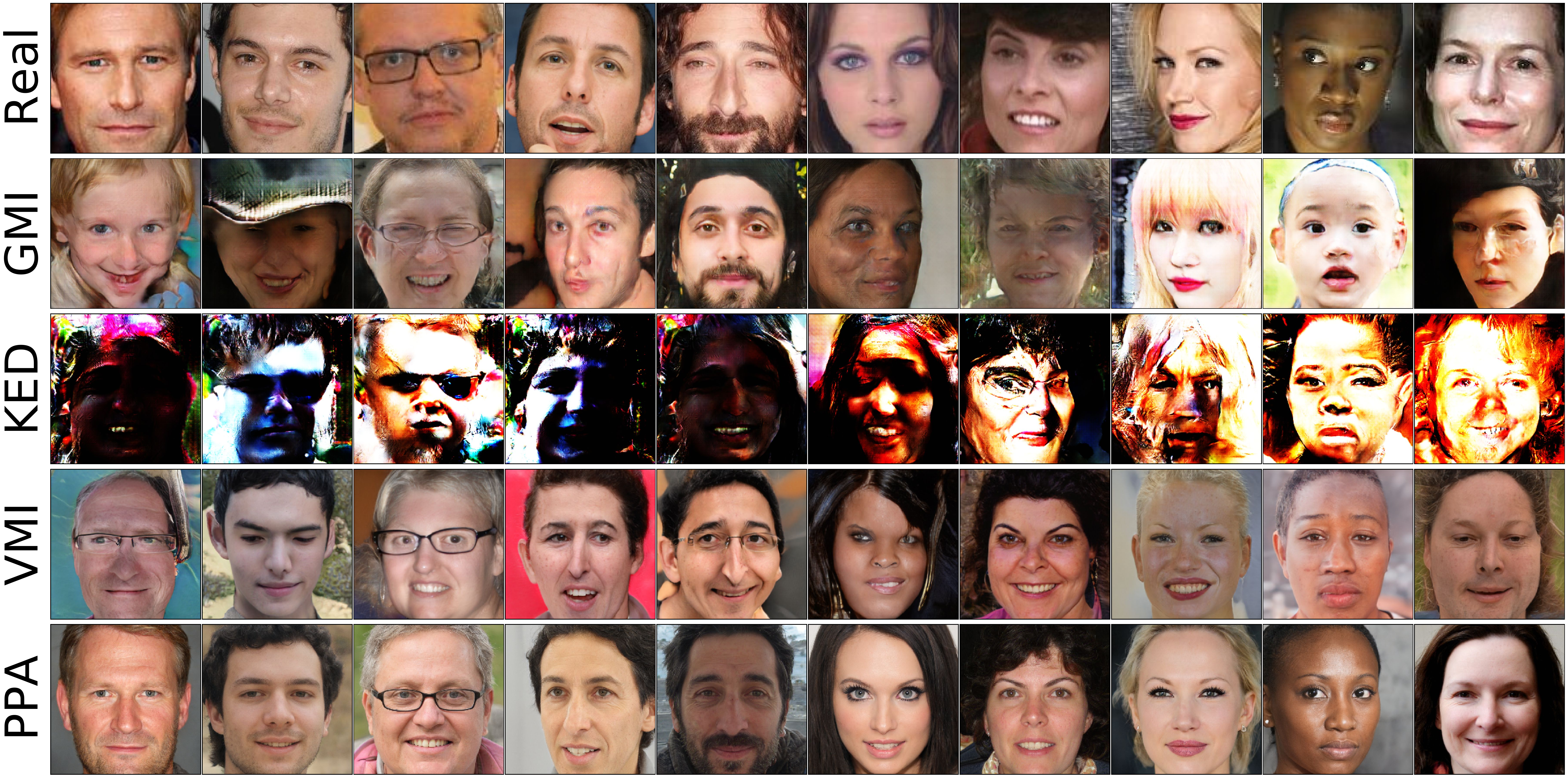}
  \vskip -0.1in
  \caption{Visual comparison of attack results against the first five actors and actresses of FaceScrub. To avoid cherry-picking, we selected the most robust samples of each attack using our transformation-based selection approach on the target model. See Fig. 9 in Appx.~\ref{app:attack_visualization} for a larger version.}
  \label{fig:visual_comparison}
  \vskip -0.1in
\end{figure}

\textbf{Comparison with Previous MIA Approaches.} We start by comparing our Plug \& Play Attacks (PPA) against the most recent work on MIAs by \citet{secret_revealer} (GMA), \citet{knowledge_mia} (KED), and \citet{variational_mia} (VMI, Gaussian approach). We carefully selected the hyperparameters by testing various configurations and hyperparameters of each attack. Detailed information on the comparison and parameter selection are available in \cref{app:attack_comparison}.

We emphasize that previous MIAs are very time-consuming: KED trains a separate GAN for each target model, and VMI even fits a separate variational model for each target class and model. Based on the same StyleGAN2 model, our approach needs about 5 minutes on a single GPU (Tesla V100-32GB) to generate 200 attack samples for arbitrary classes, whereas fitting VMI for a single class already takes about 35 minutes (Gaussian approach) and 2 hours (Flow approach), respectively. The time needed to perform KED and VMI might be feasible for low image resolutions, a small number of classes, and small target networks but increases rapidly on larger scales. Whereas all attacks need to train at least one GAN, only PPA and GMI can attack an arbitrary number of targets without additional training required. Due to the high time and resource requirements, we limited our comparison to attacking a ResNet-18 trained on FaceScrub with GANs trained on FFHQ and used our computation capacities for an extended evaluation of PPA.

\begin{table}
    \caption{Comparison of different MIA approaches against a ResNet-18 trained on FaceScrub using FFHQ for GAN training. PPA beats previous attacks on all metrics by a large margin.}
    \centering
    \vskip 0.1in
    \resizebox{\columnwidth}{!}{
    \begin{tabular}{lccccc}
    \toprule
              & $\pmb{\uparrow}$ \textbf{Acc@1}     & $\pmb{\uparrow}$ \textbf{Acc@5} & $\pmb{\downarrow}$ $\bm{\delta_{face}}$ & $\pmb{\downarrow}$ $\bm{\delta_{eval}}$  & $\pmb{\downarrow}$\textbf{FID} \\
        \midrule
        \textbf{GMI~\cite{secret_revealer}} & 13.11\% & 33.91\% & 1.2600 & 149.53 & \phantom{0}77.80 \\
        \textbf{KED~\cite{knowledge_mia}}   & 05.72\% & 13.11\% & 1.4366 & 158.03 & 207.11 \\
        \textbf{VMI~\cite{variational_mia}} & 61.63\% & 72.60\% & 0.9545 & 147.48 & \phantom{0}63.27  \\
        \textbf{PPA (Ours)} & \textbf{88.46\%} & \textbf{98.20\%} &  \textbf{0.7441} & \textbf{123.85} & \phantom{0}\textbf{41.73} \\
    \bottomrule
    \end{tabular}
    }
    \label{tab:attack_comparison}
    \vskip -0.1in
\end{table}

To date, previous MIAs have only empirically shown that their approaches work on low-resolution datasets and without significant distributional shift between the target and image prior datasets, e.g., by cropping out only the faces and removing any background information. Most experiments even assumed that the attacker had access to the exact same data distribution on which the target models were trained. However, we investigated a more realistic setting with stronger visual differences between the datasets. To facilitate the attacks, we used a slightly cropped and resized $256\times 256$ FFHQ dataset to train the GANs. PPA and VMI both rely on the same custom-trained StyleGAN2 model.

The numerical evaluation results in \cref{tab:attack_comparison} demonstrate that PPA handles distributional shifts much better than the other approaches, resulting in high attack accuracy and low feature distances. The visual comparison in \cref{fig:visual_comparison} further illustrates that previous attacks mainly fail to create realistic samples. Characteristic features are only revealed in some cases, in which the attacks seem to focus on a small subset of all features, such as glasses or beards, whereas other features are not depicted. We assume that KED, which focuses the GAN training on a specific target model, may favor the generation of fooling images in a distributional shift setting\footnote{Using a robust target model to avoid high prediction scores for fooling images provides an interesting avenue for future research.}. PPA, on the other hand, produces much more realistic-looking images and recovers most of the characteristic features. However, we note our attack also generates samples with misleading features in a few cases. While GMI and KED might fail in part due to their simple GAN architecture, VMI and PPA are performed with the same StyleGAN2 model.
\begin{figure}[t]
\centering
\includegraphics[width=\linewidth]{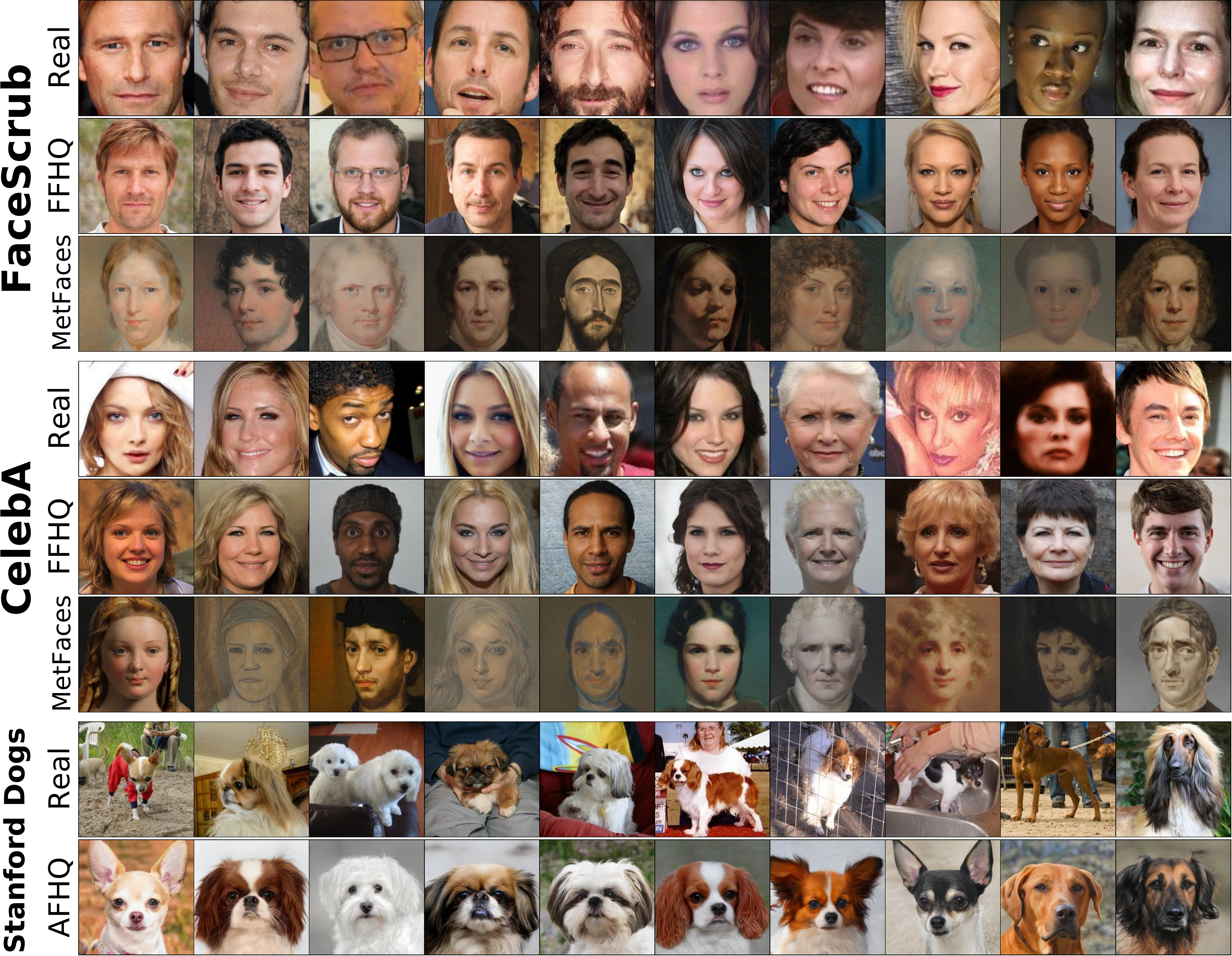}
\vskip -0.1in
\caption{Results for Plug \& Play Attacks against ResNeSt-101 models trained on FaceScrub, CelebA and Stanford Dogs, respectively. Samples were selected using our transformation-based selection approach. See \cref{fig:attack_samples_large} in Appx. \ref{app:attack_visualization} for a larger version.}
\label{fig:attack_samples}
\vskip -0.1in
\end{figure}

\textbf{Extended Evaluation.} Next, we attack a broader range of models and datasets with our PPA and publicly available, pre-trained StyleGAN2 models. We state the numerical evaluation results for attacks under various settings in \cref{tab:attack_results}. Additional results for other architectures are stated in \cref{app:add_attack_results}. All attacks targeting face classifiers using the FFHQ StyleGAN2 achieve high attack accuracy, demonstrating that the attacks create samples correctly recognized by the evaluation model. Comparing the feature distances on FaceNet to the dataset baselines -- 0.63 for FaceScrub and 0.66 for CelebA -- further underlines the successful extraction of characteristic features. \cref{fig:attack_samples} visualizes attack results against various ResNeSt-101 models. Note that the attack results have a resolution of $1024 \times 1024$, whereas the training samples were much smaller. See~\cref{app:attack_visualization} for a higher resolution and additional result visualizations.

\begin{table}[t]
    \centering
    \caption{Evaluation metrics for our PPA performed with various target datasets (first column), StyleGAN2 models (second column), and model architectures (third column).}
    \vskip 0.1in
    \resizebox{\columnwidth}{!}{
        \begin{tabular}{lllccccc}
        \toprule
            & & \textbf{Architecture} & $\pmb{\uparrow}$ \textbf{Acc@1}     & $\pmb{\uparrow}$ \textbf{Acc@5} & $\pmb{\downarrow}$ $\bm{\delta_{face}}$ & $\pmb{\downarrow}$ $\bm{\delta_{eval}}$  & $\pmb{\downarrow}$\textbf{FID} \\
            \midrule
            \parbox[t]{2mm}{\multirow{6}{*}{\rotatebox[origin=c]{90}{\textbf{FaceScrub}}}} &
            \parbox[t]{2mm}{\multirow{3}{*}{\rotatebox[origin=c]{90}{\textbf{FFHQ}}}}        & \textbf{ResNeSt-101}      & 93.95\% & 99.21\% & 0.7199 & 119.79 &  \textbf{46.30} \\
            & & \textbf{ResNet-152}       & 92.73\% & 98.91\% & 0.7163 & 123.25 & 46.69 \\
            & & \textbf{DenseNet-169}     & \textbf{95.33\%} & \textbf{99.51\%} & \textbf{0.6872} & \textbf{115.20} & 46.72 \\
            \cmidrule{2-8}
            & \parbox[t]{2mm}{\multirow{3}{*}{\rotatebox[origin=c]{90}{\textbf{\small MetFaces}}}}
              & \textbf{ResNeSt-101}      & 75.04\% & 92.74\% & 0.9787 & 137.17 & 88.66 \\
            & & \textbf{ResNet-152}       & 73.07\% & 91.95\% & 0.9660 & 139.38 & \textbf{68.54} \\
            & & \textbf{DenseNet-169}     & \textbf{79.83\%} & \textbf{94.77\%} & \textbf{0.9376} & \textbf{129.44} & 77.52 \\
            \midrule
            \parbox[t]{2mm}{\multirow{6}{*}{\rotatebox[origin=c]{90}{\textbf{CelebA}}}} &
            \parbox[t]{2mm}{\multirow{3}{*}{\rotatebox[origin=c]{90}{\textbf{FFHQ}}}}
              & \textbf{ResNeSt-101}       & \textbf{82.96\%} & \textbf{95.44\%} & 0.7506 & \textbf{299.73} & 44.04 \\
            & & \textbf{ResNet-152}        & 80.61\% & 94.58\% & \textbf{0.7362} & 312.58 & \textbf{40.43} \\
            & & \textbf{DenseNet-169}      & 73.14\% & 90.51\% & 0.7635 & 312.32 & 43.24 \\
            \cmidrule{2-8}
            & \parbox[t]{2mm}{\multirow{3}{*}{\rotatebox[origin=c]{90}{\textbf{\small MetFaces}}}}
              & \textbf{ResNeSt-101}      & 37.14\% & 62.86\% & 1.124 & \textbf{387.61} & 75.07 \\
            & & \textbf{ResNet-152}       & \textbf{39.61\%} & \textbf{64.30\%} & \textbf{1.063} & 387.81 & \textbf{74.03} \\
            & & \textbf{DenseNet-169}     & 30.90\% & 55.75\% & 1.096 & 396.81 & 81.72 \\
            \midrule
            \parbox[t]{2mm}{\multirow{3}{*}{\rotatebox[origin=c]{90}{\textbf{St. Dogs}}}} &
            \parbox[t]{2mm}{\multirow{3}{*}{\rotatebox[origin=c]{90}{\textbf{AFHQ}}}}
              & \textbf{ResNeSt-101}      & 91.90\% & 98.33\% & --    & 62.56 & 33.69 \\
            & & \textbf{ResNet-152}       & \textbf{94.98\%} & \textbf{99.57\%} & --    & \textbf{59.25} & \textbf{32.04} \\
            & & \textbf{DenseNet-169}     & 93.72\% & 99.42\% & --    & 60.03 & 32.46 \\
        \bottomrule
        \end{tabular}}
        \label{tab:attack_results}
    \vskip -0.1in
\end{table}

Using the MetFaces StyleGAN2 still achieves good evaluation results despite a large distributional shift. However, while the attacks reveal characteristic features of some classes, they fail for others. We expect this is due to the lack of diversity and features in the MetFaces dataset. We also investigate attacks against dog breed classifiers as another setting with a large distributional shift. Whereas this setting probably does not involve any privacy risk, it demonstrates the efficacy of our approach and shows that meaningful results are possible even under strong differences in the styles of dataset samples.

\textbf{Ablation Study.} We further investigated the effects of the various components of PPA and repeated the attacks against the ResNeSt-101 trained on FaceScrub using the FFHQ StyleGAN2 with individual attack components removed or adjusted. \cref{tab:ablation_study} presents the evaluation results.

\begin{table}[t]
    \centering
    \caption{Ablation study performed on a ResNeSt-101 trained on FaceScrub using the FFHQ StyleGAN2 as image prior.}
    \vskip 0.1in
    \resizebox{\columnwidth}{!}{
        \begin{tabular}{lccccc}
        \toprule
            & $\pmb{\uparrow}$ \textbf{Acc@1}     & $\pmb{\uparrow}$ \textbf{Acc@5} & $\pmb{\downarrow}$ $\bm{\delta_{face}}$ & $\pmb{\downarrow}$ $\bm{\delta_{eval}}$  & $\pmb{\downarrow}$\textbf{FID}\\
            \midrule
            \textbf{Standard PPA}      & \textbf{93.95\%} & 99.21\% & \textbf{0.7199} & 119.79 & 46.30 \\
            \textbf{CE Loss}           & \textbf{76.02\%} & 93.61\% & \textbf{0.8879} & 134.91 & 50.47 \\
            \textbf{No Center Cropping}       &  84.48\% & 9705\% & 0.8010 & 134.70 & 46.73 \\
            \textbf{Resize 168}        & 81.00\% & 95.92\% & 0.8147 & 136.86 & 46.03 \\
            \textbf{Resize 299}        & 88.65\% & 97.98\% & 0.7525 & 128.48 & 48.78 \\
            \textbf{No Random Cropping}& 92.48\% & 98.87\% & 0.7341 & 121.19 & 46.86 \\
            \textbf{No Initial Selection}  & 90.25\% & 98.43\% & 0.7601 & 126.58 & 46.55 \\
            \textbf{No Final Selection}& \textbf{80.46\%} & 93.25\% & \textbf{0.8195} & 131.13 & 48.50 \\
            \textbf{Discriminator Loss}& \textbf{79.43\%} & 95.12\% & \textbf{0.8142} & 129.78 & \textbf{45.64} \\
            \textbf{BigGAN} & \textbf{57.66\%} & 82.65\% & \textbf{1.0122} & 124.64 & 57.90 \\
        \bottomrule
        \end{tabular}
    }
    \label{tab:ablation_study}
    \vskip -0.2in
\end{table}

Using $\mathcal{L}_{CE}$ as loss function instead of $\mathcal{L}_{Poincar\acute{e}}$ led to a significant decrease of all evaluation metrics, indicating the superiority of $\mathcal{L}_{Poincar\acute{e}}$. To examine the influence of image transformation, we independently removed the center cropping or changed the resizing to 168 or 299 pixels, respectively. While the evaluation metrics decrease for all three changes, the largest performance drop is observed for resizing the images to 168 pixels and, therefore, to a smaller scale than the target model's training data. We assume it is because of missing image details due to a lower resolution. Removing the random cropping only degrades the results moderately. Similarly, the selection of the initial latent vectors to optimize also influences the results only slightly.

However, the final selection of a subset of images has a strong impact, and computing the metrics before the final selection leads to a significant drop in all evaluation metrics. For the sake of completeness, we added a discriminator loss that also degrades all metrics except the FID score, which is slightly improved.

To further investigate whether PPA is compatible with other GAN architectures, we trained a custom BigGAN~\cite{biggan} model on FFHQ as image prior. See \cref{app:biggan} for training details. Our results show that PPA still achieves good evaluation results and does not necessarily rely on StyleGAN2. However, our BigGAN model generates images of lower quality compared to StyleGAN2. We suspect that this is why the numerical results are inferior compared to the other attacks. However, fine-tuning the BigGAN training hyperparameters would most likely improve the results. We want to emphasize that these results demonstrate that PPA should, in principle, also work with pre-trained or custom-trained GANs different from {StyleGAN2}.

%% file: 6_discussion.tex
\section{Discussion, Limitations and Conclusion}\label{sec:discussion}
With Plug \& Play attacks, we introduced a new kind of MIA that is, unlike most previous attacks, independent of a particular target model and allows attacking a broad range of targets with a single GAN. We further identified various degradation factors for MIAs, including distributional shifts, vanishing gradients, and non-robust target models. To overcome vanishing gradients, we motivated the application of a Poincaré loss function instead of the traditional cross-entropy loss. Furthermore, our approach integrates random image transformations into the optimization process to improve the attack's stability and robustness. Also, we proposed an effective sample selection process to select meaningful samples from the set of attack results. In an extensive empirical evaluation, we demonstrated that our approach is the first to work reliably for distributional shifts between the target model's and image prior's training data, allowing the application of publicly available pre-trained GANs. Plug \& Play Attacks significantly outperform previous approaches, leading to state-of-the-art attack performance and the first MIA empirically suitable for high-resolution applications.

However, current MIAs, including this work, still have some limitations. All approaches rely on the availability of public data to train the image priors. We relax this assumption by including the possibility of using pre-trained models, but if no such model is available, our attack still requires a sufficiently large dataset from the targeted domain. For research purposes, data-free inversion attacks without image priors to guide the attack represent an interesting avenue. Moreover, all investigated approaches require white-box access to the target model, which is not always possible. It remains an open question whether MIAs still could produce meaningful results in a black-box setting without access to a model's gradients. The community should also think about additional metrics that include a human factor for evaluating the quality and recognition of extracted features.

We also point out that our research could be misused to attack real-world targets to infer sensitive data by illegal or unethical means. However, we believe that it is important to inform the community about the presence and feasibility of such attacks and raise awareness on the user side. Moreover, our flexible and robust approach paves the way for further analyses of factors facilitating MIAs and the development of defense strategies. We believe that these benefits outweigh any potential risks.

Apart from the adversarial setting, another interesting research direction might be the application in knowledge distillation, where we could make use of MIAs to create synthetic training samples covering the characteristic features of the different classes. From a malicious viewpoint, MIAs might also be applied in a model stealing setting, which is similar to knowledge distillation. We further imagine MIAs might improve conditional image generation by guiding the latent space optimization.

%% file: A_proofs.tex
\section{Proofs}
In this section, we state the mathematical proofs for the derivatives of the cross-entropy and the Poincaré loss functions introduced in \cref{sec:poincare_loss}.

\subsection{Derivative of Cross-Entropy Loss}\label{app:ce_derivative}
For a classification task with $C$ classes, we define the cross-entropy loss for a single sample as 
\begin{equation}
    \mathcal{L}_{CE} =- \sum_{j=1}^C t_j \log(y_j).
\end{equation}
Here, $t_c\in \{0,1\}$ denotes the one-hot encoded (ground-truth) value for class $c \in \{1,\ldots, C\}$ and $y_c$ the prediction score for the same class. To compute the prediction probabilities, a softmax function is applied to the model's output logits $o \in \mathbb{R}^C$:
\begin{equation}
    y_c = \frac{e^{o_c}}{\sum_{j=1}^C e^{o_j}}.
\end{equation}
We start by first computing the derivative of the softmax function $y_c$ with respect to the model's output logit $o_k$ using the quotient rule. In the case of $c=k$:

\begin{equation}
\begin{aligned}
    \frac{\partial y_c}{\partial o_c} 
    & = \frac{e^{o_c} \sum_{j=1}^C e^{o_j} - \left( e^{o_c} \right)^2 }{\left( \sum_{j=1}^C e^{o_j} \right)^2} \\
    & = \frac{e^{o_c}}{\sum_{j=1}^C e^{o_j}} - \left( \frac{e^{o_c}}{\sum_{j=1}^C e^{o_j}} \right)^2 \\
    & = y_c - y_c^2 = y_c(1 - y_c).
\end{aligned}
\end{equation}

For the case $c \neq k$ we additionally use the reciprocal rule:
\begin{equation}
\begin{aligned}
    \frac{\partial y_c}{\partial o_k} 
    & = e^{o_c} \frac{\partial}{\partial o_k} \left( \sum_{j=1}^C e^{o_j} \right)^{-1} \\
    & = \frac{-e^{o_c} e^{o_k}}{\left( \sum_{j=1}^C e^{o_j} \right)^{2}} \\
    & = -y_c y_k.
\end{aligned}
\end{equation}

To compute the derivative of $\mathcal{L}_{CE}$ with respect to the model's output logits, we apply the chain rule and utilize that $\sum_{j=1}^C t_j=1$ for one-class classification:

\begin{equation}
\begin{aligned}
    \frac{\partial \mathcal{L}_{CE}}{\partial o_c} 
    & = - \sum_{j=1}^C t_j \frac{\partial \log(y_j)}{\partial o_c} \\
    & = - \sum_{j=1}^C \frac{t_j}{y_j} \frac{\partial y_j}{\partial o_c} \\
    & = - \frac{t_c}{y_c} y_c (1-y_c) - \sum_{j\neq c} \frac{t_j}{y_j} (-y_j y_c) \\
    & = -t_c + y_c t_c + y_c \sum_{j \neq c} t_j \\
    & = -t_c + y_c \left( t_c + \sum_{j \neq c} t_j \right) \\
    & = y_c - t_c.
\end{aligned}
\end{equation}

\subsection{Derivative of Poincaré Loss}\label{app:poincare_derivative}
We define the Poincaré loss as the Poincaré distance between the normalized output logits $u=\frac{o}{\lVert o \rVert_1}, u\in \mathbb{R}^C$ and the adjusted one-hot encoded target vector $v\in\{0, 0.9999\}^C$ as
\begin{equation}
     \mathcal{L}_{Poincar\acute{e}}=\arcosh \left( 1 + 2\frac{\lVert u-v \rVert_2^2}{(1-\lVert u \rVert_2^2)(1-\lVert v \rVert_2^2)} \right).
\end{equation}

To make the proof easier to follow, we first define 
\begin{equation}
\begin{aligned}
\alpha=1-\lVert u \rVert_2^2,
\end{aligned}
\end{equation}

\begin{equation}
\begin{aligned}
\beta=1-\lVert v \rVert_2^2,
\end{aligned}
\end{equation}

and

\begin{equation}
\begin{aligned}
\gamma=1 + \frac{2}{\alpha \beta} \lVert u - v \rVert_2^2.
\end{aligned}
\end{equation}

We start by computing the partial derivative of $\alpha$:
\begin{equation}
\begin{aligned}
    \frac{\partial \alpha}{\partial u_c} 
    & =\frac{\partial}{\partial u_c} \left( 1 - \sum_{j=1}^C u_j^2 \right) \\
    & = -2 u_c.
\end{aligned}
\end{equation}

Next, we compute the partial derivative of $\lVert u - v \rVert_2^2$:
\begin{equation}
\begin{aligned}
    \frac{\partial \lVert u - v \rVert_2^2}{\partial u_c} 
    & =\frac{\partial}{\partial u_c} \sum_{j=1}^C (u_j - v_j)^2 \\
    & = 2(u_c - v_c).
\end{aligned}
\end{equation}

Further, the derivative of $\arcosh$ is defined as
\begin{equation}
\begin{aligned}
    \frac{\mathop{d\arcosh(x)}}{\mathop{dx}}
    & = \frac{1}{\sqrt{x^2 - 1}}.
\end{aligned}
\end{equation}

We now can derive the partial derivative of the loss using the chain and quotient rules:
\begin{equation}
\begin{aligned}
    \frac{\partial \mathcal{L}_{Poincar\acute{e}}}{\partial u_c} 
    & = \frac{\partial\arcosh(\gamma)}{\partial u_c} \\
    & = \frac{\partial\arcosh(\gamma)}{\partial \gamma} \frac{\partial\gamma}{\partial u_c} \\
    & = \frac{1}{\sqrt{\gamma^2 - 1}} \frac{\partial \gamma}{\partial u_c} \\
    & = \frac{1}{\sqrt{\gamma^2 - 1}} \frac{\partial}{\partial u_c} \left( 1 + \frac{2}{\alpha \beta} \lVert u - v \rVert_2^2 \right) \\
    & = \frac{2}{\beta \sqrt{\gamma^2 - 1}} \frac{\partial}{\partial u_c} \frac{\lVert u - v \rVert_2^2}{\alpha}  \\
    & = \frac{4}{\beta \sqrt{\gamma^2 - 1}} \frac{\alpha(u_c-v_c) + u_c \lVert u - v \rVert_2^2}{\alpha^2}.
\end{aligned}
\end{equation}

We further compute the partial derivative of $u=\frac{o}{\lVert o \lVert_1}$ and start with the case $c=k$ for the vector indices. We further make use of $\frac{d \, \lvert x\rvert}{d x} = \frac{x}{\lvert x \rvert}$ and the quotient rule and end up with:
\begin{equation}
\begin{aligned}
    \frac{\partial u_c}{\partial o_c} 
    & = \frac{\partial}{\partial o_c} \frac{o_c}{\lVert o \rVert_1} \\
    & = \frac{\sum_{j=1}^C \lvert o_j \rvert - o_c \frac{o_c}{\lvert o_c \rvert}}{\lVert o \rVert_1^2} \\
    & = \frac{\sum_{j \neq c} \lvert o_j \rvert}{\lVert o \rVert_1^2}.
\end{aligned}
\end{equation}

We then compute the partial derivative for the case $c \neq k$:
\begin{equation}
\begin{aligned}
    \frac{\partial u_c}{\partial o_k} 
    & = \frac{\partial}{\partial o_k} \frac{o_c}{\lVert o \rVert_1} \\
    & = \frac{-o_c \frac{o_k}{\lvert o_k \rvert}}{\lVert o \rVert_1^2} \\
    & = \frac{-o_c o_k}{\lvert o_k \rvert \cdot\lVert o \rVert_1^2}.
\end{aligned}
\end{equation}

We then finally derive at
\begin{equation}
\begin{aligned}
    \frac{\partial \mathcal{L}_{Poincar\acute{e}}}{\partial o_c}
    & = \sum_{j=1}^C \frac{\partial \mathcal{L}_{Poincar\acute{e}}}{\partial u_j} \frac{\partial u_j}{\partial o_c}.
\end{aligned}
\end{equation}

%% file: B_experimental_details.tex
\section{Experimental Details}\label{app:experimental_details}
Here we state the technical details of our experiments to improve reproducibility and eliminate ambiguities. 

\subsection{Hard- and Software Details}\label{app:hardware_details}
We performed all our experiments on NVIDIA DGX machines running NVIDIA DGX Server Version 5.1.0 and Ubuntu 20.04.2 LTS. The machines have 1.6TB of RAM and contain Tesla V100-SXM3-32GB-H GPUs and Intel Xeon Platinum 8174 CPUs. We further relied on CUDA 11.4, Python 3.8.10, and PyTorch 1.10.0 with Torchvision 0.11.0~\cite{pytorch} for our experiments. If not stated otherwise, we used the model architecture implementations and pre-trained ImageNet weights provided by Torchvision. We further provide a Dockerfile together with our code to make the reproduction of our results easier. In addition, all training and attack configuration files are available to reproduce the results stated in this paper.

\subsection{Evaluation Models}
We trained Inception-v3 models as evaluation models on the FaceScrub, CelebA, and cropped Stanford Dogs datasets. We used the ImageNet pre-trained models as initialization and replaced the final fully-connected layer to match the number of classes in the datasets. All datasets were split in the same way as for training the target models, using 90\% of the samples for training and 10\% for testing. We normalize all images with $\mu=\sigma=(0.5, 0.5, 0.5)$. All images were then resized so that the smaller edge of the image matches 299 pixels. We then applied random cropping with patch size varying between 85\% and 100\% of the image area and a fixed aspect ratio of $1.0$. Patches are then resized to $299\times 299$ pixels to match the expected input size of Inception~v3. We further applied color augmentations by randomly changing the brightness and contrast between $[0.8, 1.2]$, the saturation between $[0.9, 1.1]$, and the hue between $[-0.1, 0.1]$. We also horizontally flip the images with probability $p=0.5$.

All models were trained using the Adam optimizer~\cite{adam_optimizer}, with an initial learning rate of $0.001$ and $\beta=(0.9, 0.999)$. For the models trained on FaceScrub and Stanford Dogs, the learning rate was reduced by a factor of $0.1$ after 80 epochs. We trained the models for a total of 100 epochs with a batch size of 128. After training, each model's prediction accuracy is measured on the test splits. We state the test results in \cref{tab:model_acc}. We also trained an evaluation model on the uncropped Stanford Dogs dataset for comparison reasons. We did not use this model during our experiments.

For the CelebA evaluation model, we initialized the weights based on the trained FaceScrub evaluation model to increase the model's accuracy. We only replaced the final fully-connected layer to adapt for the higher number of classes and then performed the same training procedure as described above. We only adjusted the learning rate scheduler and reduced the learning rate by a factor of 0.1 after 75 and 90 epochs. By using the FaceScrub model weights as initialization, we were able to increase the test accuracy on CelebA by about three percentage points. 

We further used a pre-trained FaceNet~\cite{facenet} obtained from \href{https://github.com/timesler/facenet-pytorch}{https://github.com/timesler/facenet-pytorch} to measure the distance between training samples and attack results on the facial recognition tasks. More specifically, we used the Inception-ResNet-v1~\cite{inception_resnet} trained on VGGFace2~\cite{vggface2}. The model's declared test accuracy on LFW~\cite{LFWTech} is 99.65\%.
\begin{table}[t]
    \centering
    \vskip -0.1in
    \caption{Test accuracy of the evaluation and target models used in our experiments. The accuracy was measured on the hold-out test set after training. The train-test-splits were identical for all models trained on the same dataset.}
    \vskip 0.15in
    \begin{tabular}{llc}
    \toprule
              &                  & $\pmb{\uparrow}$ \textbf{Test Acc} \\
        \midrule
        \parbox[t]{2mm}{\multirow{14}{*}{\rotatebox[origin=c]{90}{\textbf{FaceScrub}}}}
        & \textbf{Inception-v3}           & 96.20\%   \\
        & \textbf{ResNeSt-50}           & 95.86\%   \\
        & \textbf{ResNeSt-101}          & 95.38\%   \\
        & \textbf{ResNeSt-200}          & 95.56\%   \\
        & \textbf{ResNeSt-269}          & 94.61\%   \\
        & \textbf{ResNet-18}            & 94.22\%   \\
        & \textbf{ResNet-34}            & 94.75\%   \\
        & \textbf{ResNet-50}            & 93.88\%   \\
        & \textbf{ResNet-101}           & 93.35\%   \\
        & \textbf{ResNet-152}           & 93.74\%   \\
        & \textbf{DenseNet-121}         & 95.54\%   \\  
        & \textbf{DenseNet-161}         & 94.22\%   \\
        & \textbf{DenseNet-169}         & 95.49\%   \\
        & \textbf{DenseNet-201}         & 95.17\%   \\
    \midrule
        \parbox[t]{2mm}{\multirow{4}{*}{\rotatebox[origin=c]{90}{\textbf{CelebA}}}}
        & \textbf{Inception-v3}            & 93.28\%   \\
        & \textbf{ResNeSt-101}          & 87.35\%   \\
        & \textbf{ResNet-152}           & 86.78\%   \\
        & \textbf{DenseNet-169}         & 85.39\%   \\
    \bottomrule
        \parbox[t]{2mm}{\multirow{5}{*}{\rotatebox[origin=c]{90}{\textbf{Stanf. Dogs}}}}
        & \textbf{Inception-v3 (cropped)} & 79.79\% \\
        & \textbf{Inception-v3}         & 75.17\%   \\
        & \textbf{ResNeSt-101}          & 75.07\%   \\
        & \textbf{ResNet-152}           & 71.23\%   \\
        & \textbf{DenseNet-169}         & 74.39\%   \\
    \bottomrule
    \end{tabular}
  \label{tab:model_acc}
\end{table}

\subsection{Target Models}\label{app:target_models}
Unless stated otherwise, we trained all our target models for FaceScrub, CelebA, and Stanford Dogs with the following parameters: we initialized the models with pre-trained ImageNet weights. All images are normalized with $\mu=\sigma=(0.5, 0.5, 0.5)$ and resized so that the smaller size matches 224 pixels. 

We then applied random cropping with patch size varying between 85\% and 100\% of the image area and a fixed aspect ratio of 1.0. Patches are then resized back to $224\times 224$ pixels. We further applied color augmentations by randomly changing the brightness and contrast between $[0.8, 1.2]$, the saturation between $[0.9, 1.1]$, and the hue between $[-0.1, 0.1]$. The resulting images are horizontally flipped in 50\% of the cases.

All models were trained using the Adam optimizer with an initial learning rate of $0.001$ and $\beta=(0.9, 0.999)$. We multiplied the learning rate by a factor of $0.1$ after 75 and 90 epochs. We trained the models for a total of 100 epochs with a batch size of 128. We only decreased the batch size to 96 for the DenseNet-161, 72 for the ResNeSt-200 and 64 for the ResNeSt-269. 

All models were trained on a single GPU. After training, we measured the models' prediction accuracy on the test splits. Note that we used the same train/test splits as for the evaluation models. See \cref{tab:model_acc} for evaluation accuracy results on the test set. We did not aim for achieving maximum accuracy but to apply standard training settings.

\subsection{BigGAN}\label{app:biggan}
For our ablation study, we also trained a custom BigGAN~\cite{biggan} on the FFHQ dataset. We relied on the official source code, available at \href{https://github.com/ajbrock/BigGAN-PyTorch}{https://github.com/ajbrock/BigGAN-PyTorch}. To speed up the training, we trained the model on a resized dataset with resolution $256\times 256$. We trained for 200 epochs and a batch size of 128. We further set the latent space dimension to 128 and the channel multiplier to 64 for both, the generator and discriminator. Self attention is performed at resolution $64\times 64$. We further set the learning rate for the generator to $5e-5$ and the discriminator to $2e-4$. 

We note that the resulting images are qualitatively worse than the images produced by StyleGAN2 models. However, we did not spend much time on hyperparameter tuning due to the time- and resource-consuming training. So we believe with more fine-tuning of the hyperparameters, the image quality could be improved.

\subsection{Plug \& Play Attacks}\label{app:ppa_details}
As described in the main section of this paper, we relied on the pre-trained StyleGAN2 models available at \href{https://github.com/NVlabs/stylegan2-ada-pytorch}{https://github.com/NVlabs/stylegan2-ada-pytorch}. For attacking face recognition systems, we used the models trained on FFHQ and MetFaces, respectively. Both models generate samples at $1024\times 1024$. For attacking models trained on Stanford Dogs, we used a StyleGAN2 trained on AFHQ Dog at $512\times 512$ using adaptive discriminator augmentation~\cite{Karras2020ada}. 

For the attack runs stated in this paper, we first sampled a total of 2,000 latent vectors $z \sim \mathcal{N}(0, 1)$. For attacking CelebA models, we increased the sampling number to 5,000 latent vectors to adjust for the larger number of classes. We then mapped the vectors to the intermediate latent space using $G_\mathit{mapping}$. We set the truncation to $\psi=0.5$ and the truncation cutoff to 8. The resulting intermediate latent vectors were further mapped by $G_\mathit{synthesis}$ into images. Each image was then center cropped and resized with the same parameters used during the optimization process, which we state later in this section. For each resulting image, we computed the mean prediction scores for the normal and horizontally flipped version. For each target class, we selected the 200 intermediate latent vectors $w$ as candidates that achieved the highest prediction scores for the target class.

During the optimization process, we optimized each of these 200 latent vectors for each target class. Intermediate latent vectors $w$ consist of various dimensions, and the vector at each dimension is fed into a distinct layer in $G_\mathit{synthesis}$. To shrink the optimization space, we restricted the optimization process to the first dimension of the intermediate latent vector and, consequently, optimized only a single vector with length 512 for each candidate. We then fed this single vector to each adaptive instance normalization (AdaIN) operation. Note that our attack also supports separate optimization of separate intermediate latent vectors for each AdaIN operation, but we do not make use of it in this paper.

For attacking the FaceScrub and CelebA models with the StyleGAN2 models trained on FFHQ and MetFaces, we performed center crops with size $800 \times 800$ during each forward pass and resized the resulting images to $224 \times 224$. We then applied random cropping with patch size varying between 90\% and 100\% and a fixed aspect ratio of 1.0. To optimize $w$, we used a batch size of 20 and the Adam optimizer with a learning rate of $0.005$ and $\beta=(0.1, 0.1)$. For attacking the FaceScrub models with the FFHQ StyleGAN, we optimized each batch for 50 epochs. For attacking the CelebA models or using the MetFaces StyleGAN, we increased the number of iterations to 70.

For attacking the Stanford Dog models with the StyleGAN2 trained on AFHQ dogs, we did not crop the generated images. Instead, we resized the images directly to $224\times 224$ and then applied random cropping with patch size varying between 75\% and 100\% and a fixed aspect ratio of 1.0. We used the Adam optimizer with an increased learning rate of 0.01 and $\beta=(0.1, 0.1)$ and optimized each batch for 100 iterations.

From the resulting number of 200 optimized intermediate latent vectors for each target, we then selected a total of 50 for each target. For this, we used the transformation-based selection process introduced in \cref{fig:selection_process}. For each image, we cropped 100 times a random patch with an area between $50\%$ and $90\%$ of the image's size and a ratio of width to height between $0.8$ and $1.2$. We resized the cropped image back to the image size of $224\times 224$ and computed the prediction score of $M_\mathit{target}$ for the target class. The prediction scores were then averaged across all 100 iterations. We finally selected the 50 samples with the highest average prediction scores for each target. 

The same procedure was used to select the visualized samples of our and previous MIA approaches. In all cases, we took the sample with the highest average prediction score.

\subsection{Comparison with Previous MIA Approaches}\label{app:attack_comparison}
We compare our approach with the work of \citet{secret_revealer} (GMI), \citet{knowledge_mia} (KED), and \citet{variational_mia} (VMI). KED and VMI both need to be adjusted specifically for a single target model. We trained the GANs of each attack using the FFHQ dataset. To facilitate the attacks, we center-cropped each sample with a crop size of 800. We further resized the resulting samples to size $256\times 256$ to remove the resolution gap between the image priors and the target models. By this, we made the attack setting easier compared to our setting, in which the GAN generates images at a much larger resolution and with more background information. Therefore, the investigated attacks do not need image transformations during the attack, which might have added additional instabilities. Furthermore, the time needed for training and performing the attacks is decreased significantly. For each target class, we generated 50 samples for each attack approach, matching the number of final images our PPA produced in the other experiments. We used the same evaluation models and metrics to measure the effectiveness of the different attacks as we used to evaluate the PPA results.

Unfortunately, the GAN architecture used by GMI and KED has been designed for images with a low resolution of $64\times 64$ pixels. To adjust the attack for images with a larger resolution, we customized the proposed architecture. We added two additional upsampling blocks, which consist of a transposed convolutional layer and a batch norm layer, to the generator. We kept the number of channels consistent with the original last upsampling block, which used 64 channels. The blocks were put before the final transposed convolutional layer. The resulting images have a size of $256\times 256$ pixels. We also extended the discriminator by two additional convolutional blocks, each consisting of a convolutional layer and an instance norm layer. We kept the number of feature maps consistent with the previously last convolutional block with 256 channels. Since the GAN architecture generates larger images in the $[0, 1]$ space, samples are resized to size 224 and normalized before being fed into the target or evaluation models. We set the size of the latent vector to 100 for GMI, and to 150 for KED.

\newpage
For performing GMI, we used the source code provided at \href{https://github.com/AI-secure/GMI-Attack}{https://github.com/AI-secure/GMI-Attack}. We used Adam for training the GAN with a batch size of 64, a learning rate of 0.0002, and $\beta=(0.5, 0.999)$, and trained for 280 epochs. For performing the attack, we optimized the latent vectors using SGD with a learning rate of 0.01, a momentum of 0.9, and $\lambda_i=100$. For each target class, we optimized a batch of 50 randomly sampled latent vectors for 1500 iterations and clipped the latent vectors in the space of $[-1, 1]$.

We further used the source code published at \href{https://github.com/SCccc21/Knowledge-Enriched-DMI}{https://github.com/SCccc21/Knowledge-Enriched-DMI} to perform KED. We then trained the GAN using our ResNet-18 trained on FaceScrub with 530 classes to produce soft-labels. We further used a learning rate of 0.004, a batch size of 32, and the Adam optimizer with $\beta=(0.5, 0.999)$. We set the dimension of the latent vector to 150. We also experimented with a size of 100 and other learning rates (0.02, 0.0002) but it led to inferior evaluation performance. The full architecture was trained for 280 epochs and the generator was updated every five steps. After training, we set $\lambda_i = 100$ and optimized each $\mu$ and $\sigma$ with a learning rate of 0.02 for 2500 iterations to perform the attack. We also tried other combinations of the learning rate (0.01, 0.005, 0.002) and the number of iterations (500, 1500, 2500) but were not able to improve the results. For each target class, we repeated the attack ten times and drew after each run five random samples of $\epsilon$ and generated the corresponding images $G(\sigma \epsilon + \mu)$.

To perform VMI, we relied on the source code available at \href{https://github.com/wangkua1/vmi}{https://github.com/wangkua1/vmi}. Since the attack can also make use of the StyleGAN2 architecture, we trained a new StyleGAN2 using the official source code from \href{https://github.com/NVlabs/stylegan2-ada-pytorch}{https://github.com/NVlabs/stylegan2-ada-pytorch}. We trained the model using the provided configuration \textit{paper256} on 4 GPUs and for a total of 25,000k images. The trained model achieves an FID score of 3.60, measured against 50,000 real samples. 

Note that it is also possible to use the pre-trained StyleGAN2 we used for PPA. However, we decided to train and use a smaller StyleGAN2 to avoid attack distortions due to image transformations during the attack. Furthermore, the time and computational resources needed to perform the attack increased significantly for the larger StyleGAN2 model, making the attack hardly feasible in the scope of this work. 

For the VMI attack itself, we used the Gaussian approach, which is much faster than the Flow approach and promises only slightly worse results compared to Flow. We trained each GMM model to learn the variational distribution of a class using a batch size of 64 images and a learning rate of 0.0005. We further trained each model for 20 epochs. We kept the other parameters at their default values, as stated in the official source code.

For a fair comparison, we performed our PPA using the same StyleGAN2 model that was used for performing VMI. We kept almost all attack parameters the same as for our other attacks targeting FaceScrub models. Only the center crop transformation is removed since the StyleGAN already produces partly cropped images.  

\subsection{Ablation Study}
We performed the ablation study using the same ResNeSt-101 trained on FaceScrub as the target model. Also, the StyleGAN2 FFHQ model was used as the image prior in all cases. We further point out that we only changed one part of the attack, keeping all other parts as for the Standard PPA. In the case of the cross-entropy loss, we tried various learning rates (0.001, 0.005, 0.01) and selected the best results, which were achieved by setting the learning rate to 0.01. For the setting without an initial selection, we also sampled 2,000 random latent vectors but then picked the 200 candidates for each target class randomly from the set instead of using our initial selection procedure. For the setting without a final selection, we skipped the selection of a final subset of results and computed the metrics on all optimized vectors. For the discriminator loss, we used the pre-trained StyleGAN2 discriminator and weight the loss with 0.1. Lowering the weight improved the results, which is not surprising.

For performing PPA with BigGAN, we increased the number of optimization iterations to 100 and set the learning rate to 0.05. We further remove the center cropping since the generated images are already of size $256\times 256$. All other parameters are set the same as for the attacks using StyleGAN2 Image priors to show that not much parameter tuning is needed if the image prior changes. See \cref{app:ppa_details} for more details.

\subsection{Fréchet Inception Distance}\label{app:fid_score}
With $\mu_i$ and $\Sigma_i$ being the mean and covariance matrix of the feature vectors extracted from $X_\mathit{target}$ and $\hat{X}_{attack}$, respectively, the FID is computed by
\begin{equation}
\begin{aligned}
    FID &= \lVert \mu_\mathit{target} - \mu_{attack} \rVert_2^2  \\
        & + Tr \left( \Sigma_\mathit{target} + \Sigma_{attack} - 2(\Sigma_\mathit{target} \Sigma_{attack})^{0.5}) \right).
\end{aligned}
\end{equation}

Previous work only computed the FID score on samples correctly classified by the evaluation model. So even for an MIA that produces only a few correctly classified samples and a large number of unrelated and falsely classified samples, the FID score might still state good results. We further argue that a strong MIA does not necessarily need to generate samples close to the training distribution and, consequently, an attack might produce privacy leaking results that result in a low FID score. For the face recognition example, an attacker probably is not interested in retrieving the full bandwidth of the training distribution, e.g., samples showing a person of various ages or under different weather and lighting conditions. Even if the generated images do not represent the training data distribution, the identity of other sensitive information might correctly be inferred from the results.

\subsection{Experimental Details for \cref{fig:mia_problems}}\label{app:mia_problems_details}
In the following, we describe the parameters used to create the attack results depicted in \cref{fig:mia_problems}, starting from the left. We used our ResNeSt-101 trained on FaceScrub as target model. The first image belongs to the first identity from the FaceScrub training set. The second image visualizes a distributional shift that was created by performing our attack without any image transformations except resizing. We resized the images to $800\times 800$, a significantly larger resolution than our target model was trained on. The third picture showing a local minimum was created using a cross-entropy loss instead of the Poincaré loss for the optimization process. The fourth picture was created using a higher learning rate of 0.1 in combination with Adam. For the third and fourth images, we further did not use random cropping and only performed center cropping and resizing as in the other attacks. We manually selected the images from a set of eight images for each attack setting.

\subsection{Experimental Details for \cref{fig:loss_comparison}}\label{app:loss_comparison_details}
We here state the details for creating \cref{fig:loss_comparison} on a trained ResNeSt-101. We first sampled 100 random intermediate latent vectors $w$ and used the single latent vectors that achieved the highest initial prediction scores for the target class. To avoid cherry-picking, we used the first five target classes for the actors and actresses in the FaceScrub dataset corresponding to indices 0-4 and 265-269. Consequently, we optimized a total of ten samples. 

We performed only a center crop to size 800 and resized the images to size $224\times 224$ during optimization to avoid random influences. Adam with a learning rate of $0.005$ and $\beta=(0.1, 0.1)$ was used for optimization. The computed gradient norms were rescaled by dividing them through the norm values of the first epoch to make both approaches visually comparable. 

Increasing the learning rate for $\mathcal{L}_{CE}$ did either not increase the induced changes or led to unrealistic changes in other candidate images. While larger learning rates of ($0.05, 0.1$) might help poor initial local minima, many results overshoot the optimal solutions and end up in unrealistic images. So compared to $\mathcal{L}_{Poincar\acute{e}}$, a more careful selection of optimization parameters individually for distinct initial latent vectors is needed to achieve good results. The same conclusions can be drawn from experimenting using SGD instead of Adam and learning rates $\in \{0.01, 0.1, 1.0\}$.

%% file: C_datasets.tex
\section{Datasets}\label{app:dataset}

\begin{figure}[t]
\centering
\includegraphics[width=\linewidth]{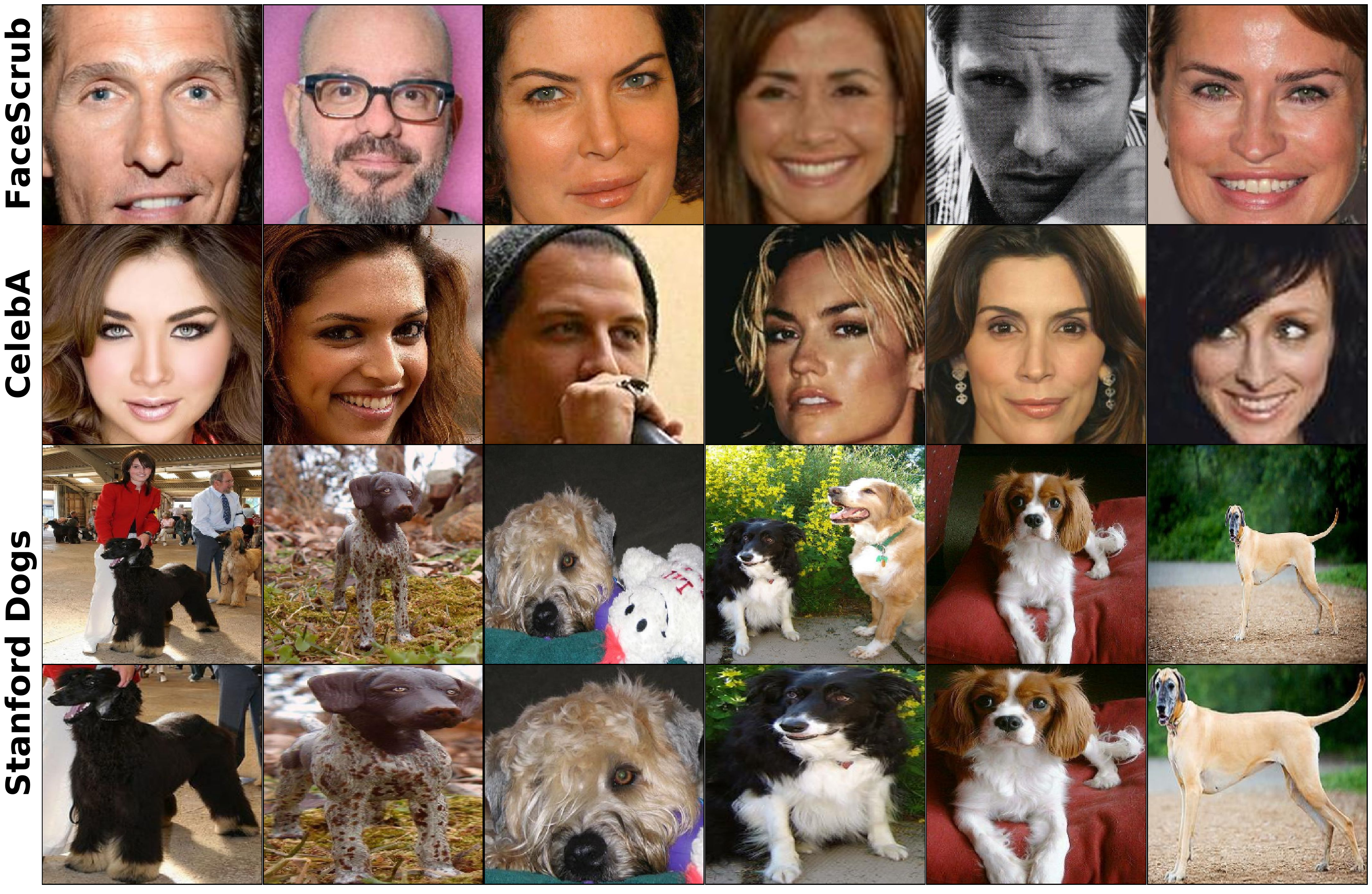}
\vskip -0.1in
\caption{Randomly selected samples from the datasets used to train the target and evaluation models. For the Stanford Dogs dataset, both uncropped and cropped samples are depicted.}
\vskip 0.1in
\label{fig:dataset_samples}

\end{figure}

\begin{figure}[t]
\centering
\includegraphics[width=\linewidth]{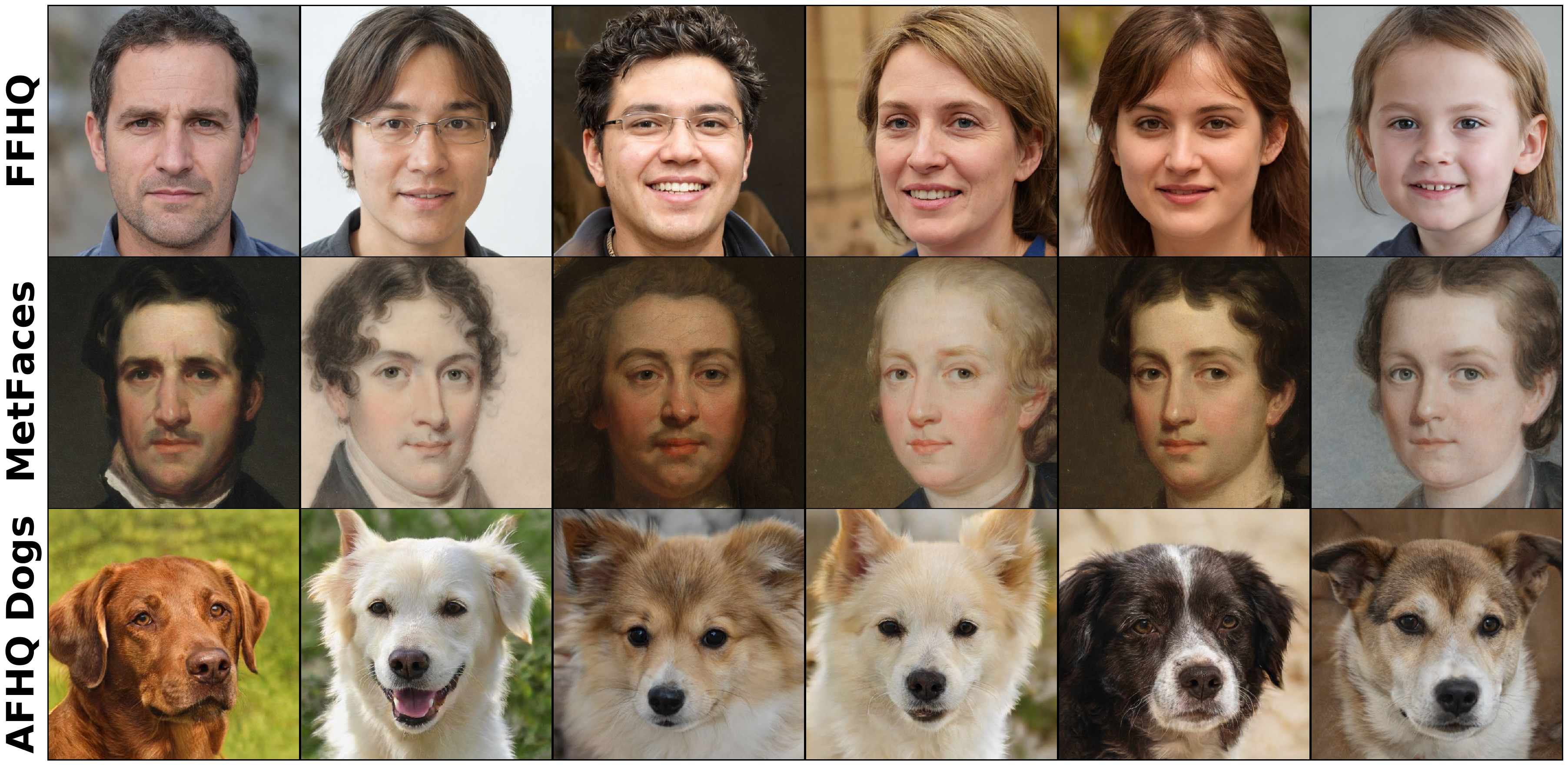}
\vskip -0.1in
\caption{Randomly generated samples from the StyleGAN2 models used to attack the target models.}
\label{fig:stylegan_samples}
\end{figure}

We split each training dataset into 90\% training data and 10\% test data. The splits are identical for all target and evaluation models. The test data is only used to evaluate the models' prediction accuracy. We further want to point out that neither the StyleGAN2 models nor the attacks have access to the target models' training data. The datasets are only used to evaluate the attack results after the attacks are finished, and the final generated images were selected. We visualize samples from the different target training sets in \cref{fig:dataset_samples}. We further sampled images using the three StyleGAN2 models used for the attacks. The samples are depicted in \cref{fig:stylegan_samples}.

We also computed the FID scores between the different datasets used to train the StyleGAN2 and target models to quantify the visual differences between the datasets and build a baseline for the attacks. Results are stated in \cref{tab:dataset_fid}. We further measured the mean feature distances between the samples of the same class in our datasets. 

We then averaged the results across all classes and stated the results together with the standard deviation in \cref{tab:dataset_dist}. Note that we did not compute the minimum distances between samples to take into account that FaceScrub and CelebA contain samples from the same class that are visually identical and only vary in resolution or cropped image parts. Also, some samples are the horizontally flipped version of another sample. 

\textbf{FaceScrub.} FaceScrub provides cropped face images of 530 identities and is published under the CC BY-NC-ND 3.0 license\footnote{\href{https://creativecommons.org/licenses/by-nc-nd/3.0/}{https://creativecommons.org/licenses/by-nc-nd/3.0/}}. Originally, FaceScrub provided 106,863 samples, but since only links instead of images are provided, we were only able to obtain 37,878 samples. In our train/test split, it resulted in a total of 34,090 training and 3,788 test samples for FaceScrub. We want to point out that a significant share of FaceScrub samples is mislabelled or of low resolution, which makes learning to classify the images even harder. Also, images of a specific target might strongly vary in style, e.g., showing a person at different ages, with different hair colors, facial accessories, etc. The FaceScrub dataset is available at \href{http://vintage.winklerbros.net/facescrub.html}{http://vintage.winklerbros.net/facescrub.html}.

\textbf{CelebA.} CelebA also contains facial images of celebrities and is available for non-commercial research purposes only. To increase the quality of the CelebA samples, we create a new dataset of cropped and aligned images using the HD CelebA Cropper, available at \href{https://github.com/LynnHo/HD-CelebA-Cropper}{https://github.com/LynnHo/HD-CelebA-Cropper}. We cropped the images using a face factor of 0.65 and resized them to $224\times 224$ using bicubic interpolation. The other parameters were left at default. We then took the 1,000 identities with the most number of samples out of 10,177 available identities. This leaves us with 27,034 training and 3,004 test samples. The total dataset consists of 202,599 images. For comparison, the official CelebA dataset of aligned faces only provides images with a size of $218 \times 178$. The CelebA dataset is available at \href{https://mmlab.ie.cuhk.edu.hk/projects/CelebA.html}{https://mmlab.ie.cuhk.edu.hk/projects/CelebA.html}.
\begin{table}[t]
 \caption{We computed the FID scores between the various datasets used to train the StyleGAN2 prior and the target models. A larger score indicates a larger divergence between the datasets. }
 \vskip 0.1in
\small
    \centering
    \begin{tabular}{llc}
    \toprule
    \textbf{Dataset 1} & \textbf{Dataset 2} & \textbf{FID} \\
    \midrule
        FFHQ &  FaceScrub & 77.90 \\
        FFHQ & CelebA & 59.48 \\
        FaceScrub & CelebA & 21.51 \\
        \midrule
        MetFaces & FaceScrub & 104.33 \\
        MetFaces & CelebA & 93.64 \\
        \midrule
        AFHQ Dogs & Stanford Dogs & 43.99 \\
        AFHQ Dogs & Stanford Dogs Cropped & 37.23 \\
        Stanford Dogs & Stanford Dogs Cropped & 5.48 \\
    \bottomrule
    \end{tabular}
  \label{tab:dataset_fid}
  \vskip -0.1in
\end{table}

\begin{table}[t]
 \caption{Mean inner-class distances on the training data measured on FaceNet and the evaluation Inception-v3 models. We computed for each training sample and class the mean distance to all other samples from the same class. The standard deviation is measured between different classes.}
  \vskip 0.1in
\small
    \centering
    \resizebox{\columnwidth}{!}{
    \begin{tabular}{lcc}
    \toprule
    \textbf{Dataset} & $\bm{\delta_{face}}$ & $\bm{\delta_{eval}}$ \\
    \midrule
        \textbf{FaceScrub} & $0.6303 \pm 0.13$ & $168.26 \pm  21.04$ \\
        \textbf{CelebA} & $0.6610 \pm 0.19$ & $315.20 \pm 58.30$ \\
        \textbf{Stanford Dogs} & -- & $144.26 \pm 22.66$ \\
        \textbf{Stanford Dogs Cropped} & -- & $125.76 \pm 21.94$ \\
    \bottomrule
    \end{tabular}
    }
  \label{tab:dataset_dist}
  \vskip -0.1in
\end{table}

\textbf{Stanford Dogs.} The dataset is built upon ImageNet, which is available for non-commercial research purposes only, and provides 20,580 images of 120 dog breeds. Our training and test splits consist of 18,522 and 2,058 samples, respectively. The images have no fixed size ratio and vary highly in style and content. Some images only contain a single dog, other samples picture more than one dog from the same breed. 

A significant share of images also shows humans or large parts of the background. This makes learning this dataset quite hard. While we trained the target models on the raw samples, we used a cropped version of the images for training the evaluation model. We used the officially provided bounding box annotations to crop each sample before applying additional transformations. We expect the evaluation model to make more precise predictions on the attack results, which were created using a StyleGAN2 trained on AFHQ Dogs. 

If the evaluation model were trained on the raw images, an evaluation bias might have been introduced since the trained model might not be able to correctly classify generated samples due to the strong distributional shift between AFHQ Dogs and the uncropped Stanford Dog images. The Stanford Dogs dataset is available at \href{http://vision.stanford.edu/aditya86/ImageNetDogs}{http://vision.stanford.edu/aditya86/ImageNetDogs}.

\textbf{Flickr-Faces-HQ (FFHQ).} The dataset contains 70,000 high-quality images of human faces, crawled from Flickr, and is published under the CC BY-NC-SA 4.0 license\footnote{\href{https://creativecommons.org/licenses/by-nc-sa/4.0/}{https://creativecommons.org/licenses/by-nc-sa/4.0/}}. Each image has a size of $1024\times 1024$. The authors aimed to provide a large variation in terms of age, ethnicity, and image background. Many samples also contain various accessories, e.g., eyeglasses, sunglasses, earrings, etc. The overall image quality is much higher and more consistent compared to FaceScrub and CelebA. Moreover, not only the faces but the entire heads and additional background information is depicted by the samples. The FFHQ dataset and the licenses for individual images are available at \href{https://github.com/NVlabs/ffhq-dataset}{https://github.com/NVlabs/ffhq-dataset}.

\newpage
\textbf{MetFaces.} The dataset contains 1,336 high-quality images of drawn human faces from art and is published under the CC BY-NC 2.0 license\footnote{\href{https://creativecommons.org/licenses/by-nc/2.0/}{https://creativecommons.org/licenses/by-nc/2.0/}}.  Various art styles are covered, and the samples differ significantly from each other. The dataset has a large bias towards people with light skin color and only contains a few samples of faces with darker skin. The MetFaces dataset the licenses for individual images are available at \href{https://github.com/NVlabs/metfaces-dataset}{https://github.com/NVlabs/metfaces-dataset}.

\textbf{Animal Faces-HQ (AFHQ).} The full AFHQ dataset consists of 16,130 high-resolution images at a resolution of $512\times512$. The dataset contains samples from cats, dogs, and wildlife animals, and is published under the CC BY-NC 4.0 license\footnote{\href{https://creativecommons.org/licenses/by-nc/4.0/}{https://creativecommons.org/licenses/by-nc/4.0/}}. The StyleGAN2 we used to attack models trained on Stanford Dogs models was trained on the AFHQ Dog split containing 4,739 training and 500 validation samples. The dog faces are centered and the image style is far more consistent than for the Stanford Dogs dataset. The distributional shift between both datasets is significantly larger than between the various facial image datasets. The AFHQ (Dog) dataset is available at \href{https://github.com/clovaai/stargan-v2}{https://github.com/clovaai/stargan-v2}.

%% file: D_add_results.tex
\section{Additional Experimental Results}\label{app:add_details}
In this section, we state additional attack results that did not fit into the main part of the paper.

\subsection{Attacks against FaceScrub Targets}\label{app:add_attack_results}
Apart from the results stated in the main part for attacking models trained on FaceScrub, we here state additional results for attacking a broader range of model architectures. Results are state in \cref{tab:facescrub_add_results}. All models were trained with the same training procedure described in \cref{app:target_models}. The performed attacks followed the parameters stated in \cref{app:ppa_details}.

\begin{table}[h]
    \centering
    \caption{Evaluation results for attacking a broader range of targets trained on FaceScrub.}
    \vskip 0.1in
    \resizebox{\columnwidth}{!}{
    \begin{tabular}{llccccc}
    \toprule
              & $\pmb{\uparrow}$ \textbf{Acc@1}     & $\pmb{\uparrow}$ \textbf{Acc@5} & $\pmb{\downarrow}$ $\bm{\delta_{face}}$ & $\pmb{\downarrow}$ $\bm{\delta_{eval}}$  & $\pmb{\downarrow}$\textbf{FID}\\
        \midrule
        \textbf{ResNeSt-50}  & 93.03\% & 98.82\% & 0.7270 & 121.98 & 46.92 \\
        \textbf{ResNeSt-101} & \textbf{93.95\%} & \textbf{99.21\% }& \textbf{0.7199} & \textbf{119.79} & 46.30 \\
        \textbf{ResNeSt-200} & 90.82\% & 98.51\% & 0.7265 & 124.24 & 47.23 \\
        \textbf{ResNeSt-269} & 89.98\% & 98.34\% & 0.7217 & 125.24 & \textbf{45.35} \\
        \midrule
        \textbf{ResNet-18}   & \textbf{95.48\%} & \textbf{99.57\%} & 0.6867 & \textbf{112.24} & \textbf{45.82} \\
        \textbf{ResNet-34}   & 95.06\% & 99.42\% & \textbf{0.6853} & 112.37 & 48.18 \\
        \textbf{ResNet-50}   & 89.56\% & 98.12\% & 0.7320 & 122.82 & 47.90 \\
        \textbf{ResNet-101}  & 91.37\% & 98.77\% & 0.7169 & 125.25 & 46.29 \\
        \textbf{ResNet-152}  & 92.73\% & 98.91\% & 0.7163 & 123.25 & 46.69 \\
        \midrule
        \textbf{DenseNet-121}& 95.13\% & \textbf{99.50\% }& \textbf{0.6841} & 116.14 & 46.92 \\
        \textbf{DenseNet-161}& 91.49\% & 98.77\% & 0.7083 & 123.41 & 46.92 \\
        \textbf{DenseNet-169}& \textbf{95.33\%} & \textbf{99.51\%} & 0.6872 & \textbf{115.20} & \textbf{46.72} \\        \textbf{DenseNet-201}& 93.81\% & 99.20\% & 0.7081 & 119.17 & 46.98 \\
    \bottomrule
    \end{tabular}
    }
  \label{tab:facescrub_add_results}
\end{table}

\subsection{Additional Evaluation Metrics}\label{app:add_metric_results}
We further followed \citet{variational_mia} and computed the improved precision and recall~\cite{improved_precision} together with the density and coverage~\cite{coverage_density} on a per-class basis to evaluate the sample diversity. More specifically, we used the Inception-v3 model that was also used to compute the FID score to compute the four metrics.

We state the results for comparing our PPA against previous approaches in \cref{tab:add_attack_results_comparison}. The same metrics were also measured for PPA in various settings and the ablation study. See \cref{tab:add_attack_results_improved} for the additional settings and \cref{tab:ablation_study_add_metrics} for the ablation study.

\begin{table}[ht]
    \centering
    \vskip -0.1in
     \caption{Improved precision and recall, density, and coverage metrics for our and previous MIAs performed against a ResNet-18 trained on FaceScrub.}
    \vskip 0.1in
    \resizebox{\columnwidth}{!}{
    \begin{tabular}{lcccc}
    \toprule
    & $\pmb{\uparrow}$ \textbf{Precision}     & $\pmb{\uparrow}$ \textbf{Recall} & $\pmb{\uparrow}$ \textbf{Density} & $\pmb{\uparrow}$ \textbf{Coverage}  \\
        \midrule
        \textbf{GMI~\cite{secret_revealer}} & 0.0133 & \textbf{0.3232} & 0.0319 & 0.0520 \\
        \textbf{KED~\cite{knowledge_mia}}   & 0.0011 & 0.0000 & 0.0027 & 0.0006 \\
        \textbf{VMI~\cite{variational_mia}} & 0.0246 & 0.0273 & 0.0370 & 0.0454 \\
        \textbf{PPA (Ours)}                 & \textbf{0.2837} & 0.0496 & \textbf{0.2711} & \textbf{0.2125} \\
    \bottomrule
    \end{tabular}
    }
  \label{tab:add_attack_results_comparison}
\end{table}

\begin{table}[ht]
    \centering
    \vskip -0.1in
     \caption{Improved precision and recall, density, and coverage metrics for performing our PPA in various settings.}
     \vskip 0.1in
    \resizebox{\columnwidth}{!}{
    \begin{tabular}{lllcccc}
    \toprule
              & & & $\pmb{\uparrow}$ \textbf{Precision}     & $\pmb{\uparrow}$ \textbf{Recall} & $\pmb{\uparrow}$ \textbf{Density} & $\pmb{\uparrow}$ \textbf{Coverage}  \\
        \midrule
        \parbox[t]{2mm}{\multirow{16}{*}{\rotatebox[origin=c]{90}{\textbf{FaceScrub}}}} &
        \parbox[t]{2mm}{\multirow{13}{*}{\rotatebox[origin=c]{90}{\textbf{FFHQ}}}}    
          & \textbf{ResNeSt-50}   & 0.1417 & 0.0100 & 0.1652 & 0.1164 \\
        & & \textbf{ResNeSt-101}  & 0.1556 & 0.0059 & 0.1767 & 0.1485 \\
        & & \textbf{ResNeSt-200}  & \textbf{0.1563} & \textbf{0.0277} & \textbf{0.2382} & \textbf{0.2131} \\
        & & \textbf{ResNeSt-269}  & 0.1485 & 0.0192 & 0.2335 & 0.1980 \\
        \cmidrule{3-7}
        & & \textbf{ResNet-18}    & \textbf{0.2211} & 0.0020 & \textbf{0.2733} & 0.1567 \\
        & & \textbf{ResNet-34}    & 0.1402 & 0.0059 & 0.1642 & 0.1159 \\
        & & \textbf{ResNet-50}    & 0.1286 & 0.0102 & 0.1662 & 0.1076 \\
        & & \textbf{ResNet-101}   & 0.1888 & \textbf{0.0195} & 0.2714 & \textbf{0.1953} \\
        & & \textbf{ResNet-152}   & 0.1533 & 0.0020 & 0.2103 & 0.0950 \\
        \cmidrule{3-7}
        & & \textbf{DenseNet-121} & 0.1879 & 0.0063 & 0.2129 & 0.1337 \\
        & & \textbf{DenseNet-161} & 0.1773 & \textbf{0.0373} & 0.2174 & \textbf{0.1349} \\
        & & \textbf{DenseNet-169} & \textbf{0.2279} & 0.0015 & \textbf{0.2562} & 0.1334 \\
        & & \textbf{DenseNet-201} & 0.1472 & 0.0371 & 0.2027 & 0.1340 \\
        \cmidrule{2-7}
        & \parbox[t]{2mm}{\multirow{3}{*}{\rotatebox[origin=c]{90}{\textbf{\small MetFaces}}}}    
        & \textbf{ResNeSt-101}    & 0.0694 & 0.0002 & 0.1195 & 0.0688 \\
        & & \textbf{ResNet-152}   & 0.0626 & 0.0005 & 0.0970 & 0.0704 \\
        & & \textbf{DenseNet-169} & \textbf{0.1444} & \textbf{0.0533} & \textbf{0.1797} & \textbf{0.0816} \\
        \midrule
        \parbox[t]{2mm}{\multirow{6}{*}{\rotatebox[origin=c]{90}{\textbf{CelebA}}}} &
        \parbox[t]{2mm}{\multirow{3}{*}{\rotatebox[origin=c]{90}{\textbf{FFHQ}}}}    
          & \textbf{ResNeSt-101}  & 0.2650 & 0.0136 & \textbf{0.8547} & 0.3624  \\
        & & \textbf{ResNet-152}   & \textbf{0.3231} & 0.0269 & 0.7984 & 0.2805  \\
        & & \textbf{DenseNet-169} & 0.2049 & \textbf{0.0495} & 0.6811 & \textbf{0.3866}  \\
        \cmidrule{2-7}
        & \parbox[t]{2mm}{\multirow{3}{*}{\rotatebox[origin=c]{90}{\textbf{\small MetFaces}}}}  
          & \textbf{ResNeSt-101}  & 0.0438 & 0.0008 & 0.2371 & \textbf{0.1083}  \\
        & & \textbf{ResNet-152}   & 0.0497 & 0.0002 & \textbf{0.2386} & 0.0843  \\
        & & \textbf{DenseNet-169} & \textbf{0.0529} & \textbf{0.0110} & 0.2382 & 0.0935 \\
        \midrule 
        \parbox[t]{2mm}{\multirow{3}{*}{\rotatebox[origin=c]{90}{\textbf{St. Dogs}}}} &
        \parbox[t]{2mm}{\multirow{3}{*}{\rotatebox[origin=c]{90}{\textbf{AFHQ}}}}    
          & \textbf{ResNeSt-101}  & \textbf{0.4723} & 0.0121 & \textbf{0.2567} & 0.1617 \\
        & & \textbf{ResNet-152}   & 0.3528 & 0.0085 & 0.2168 & 0.1846 \\
        & & \textbf{DenseNet-169} & 0.4196 & \textbf{0.0135} & 0.2391 & \textbf{0.1876} \\
    \bottomrule
    \end{tabular}
    }
  \label{tab:add_attack_results_improved}
\end{table}

\begin{table}[H]
    \centering
    \caption{Improved precision and recall, density, and coverage metrics for our ablation study performed on a ResNeSt-101 trained on FaceScrub.}
    \vskip 0.1in
    \resizebox{\columnwidth}{!}{
    \begin{tabular}{lcccc}
    \toprule
        & $\pmb{\uparrow}$ \textbf{Precision}     & $\pmb{\uparrow}$ \textbf{Recall} & $\pmb{\uparrow}$ \textbf{Density} & $\pmb{\uparrow}$ \textbf{Coverage}  \\
        \midrule
        \textbf{Standard PPA}       & 0.1556 & 0.0059 & 0.1767 & 0.1485 \\
        \textbf{CE Loss}            & 0.1369 & 0.0928 & 0.1608 & 0.1337 \\
        \textbf{No Center Cropping} & 0.1564 & 0.0056 & 0.1828 & 0.1671 \\
        \textbf{No Random Cropping} & 0.1416 & 0.0055 & 0.1901 & 0.1557 \\
        \textbf{Resize 168}         & 0.1776 & 0.0011 & 0.1954 & 0.1231 \\
        \textbf{Resize 299}         & 0.1928 & 0.0015 & 0.2062 & 0.1919 \\
        \textbf{No Initial Selection}&0.1946 & 0.0194 & 0.2957 & 0.1512 \\
        \textbf{No Final Selection} & 0.1518 & 0.0185 & 0.7556 & 0.2455 \\
        \textbf{Discriminator Loss} & 0.1497 & 0.0224 & 0.1733 & 0.1317 \\ 
        \textbf{BigGAN}             & 0.0486 & 0.1365 & 0.0683 & 0.0791 \\
    \bottomrule
    \end{tabular}
    \label{tab:ablation_study_add_metrics}
    }
\end{table}

\newpage
\subsection{Visualization of Attack Results}\label{app:attack_visualization}
\begin{figure*}[ht]
\centering
\includegraphics[width=\linewidth]{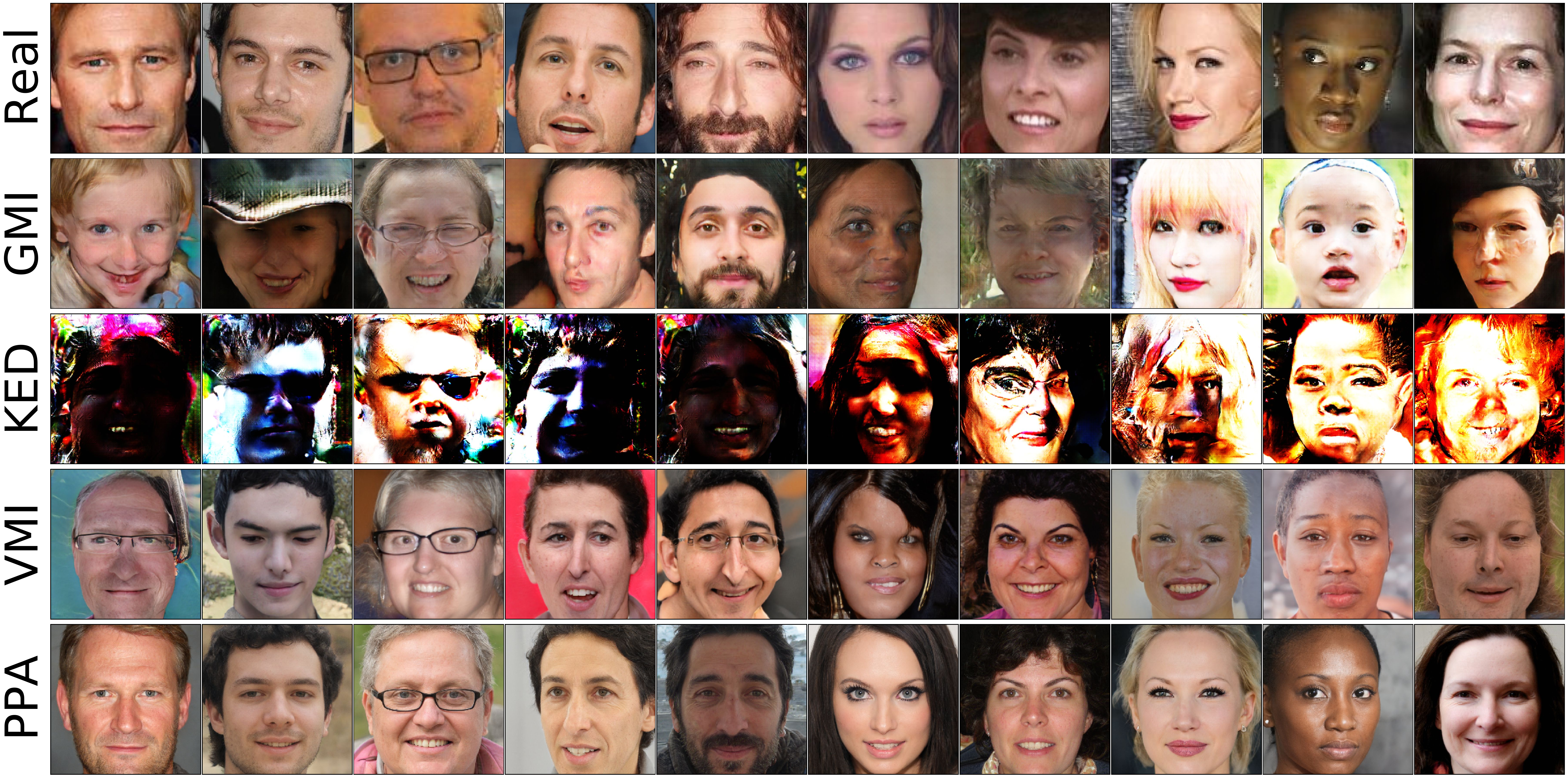}
\caption{Visual comparison of attack results against the first five actors and actresses of FaceScrub. To avoid cherry-picking, we selected the most robust samples of each attack using our transformation-based selection approach.}
\label{fig:visual_comparison_large}
\end{figure*}

To enable a better visual analysis of the attack results from our and previous approaches, we plotted \cref{fig:visual_comparison} from the main paper in a larger resolution in \cref{fig:visual_comparison_large}.

\begin{figure*}[ht]
\centering
\includegraphics[width=\linewidth]{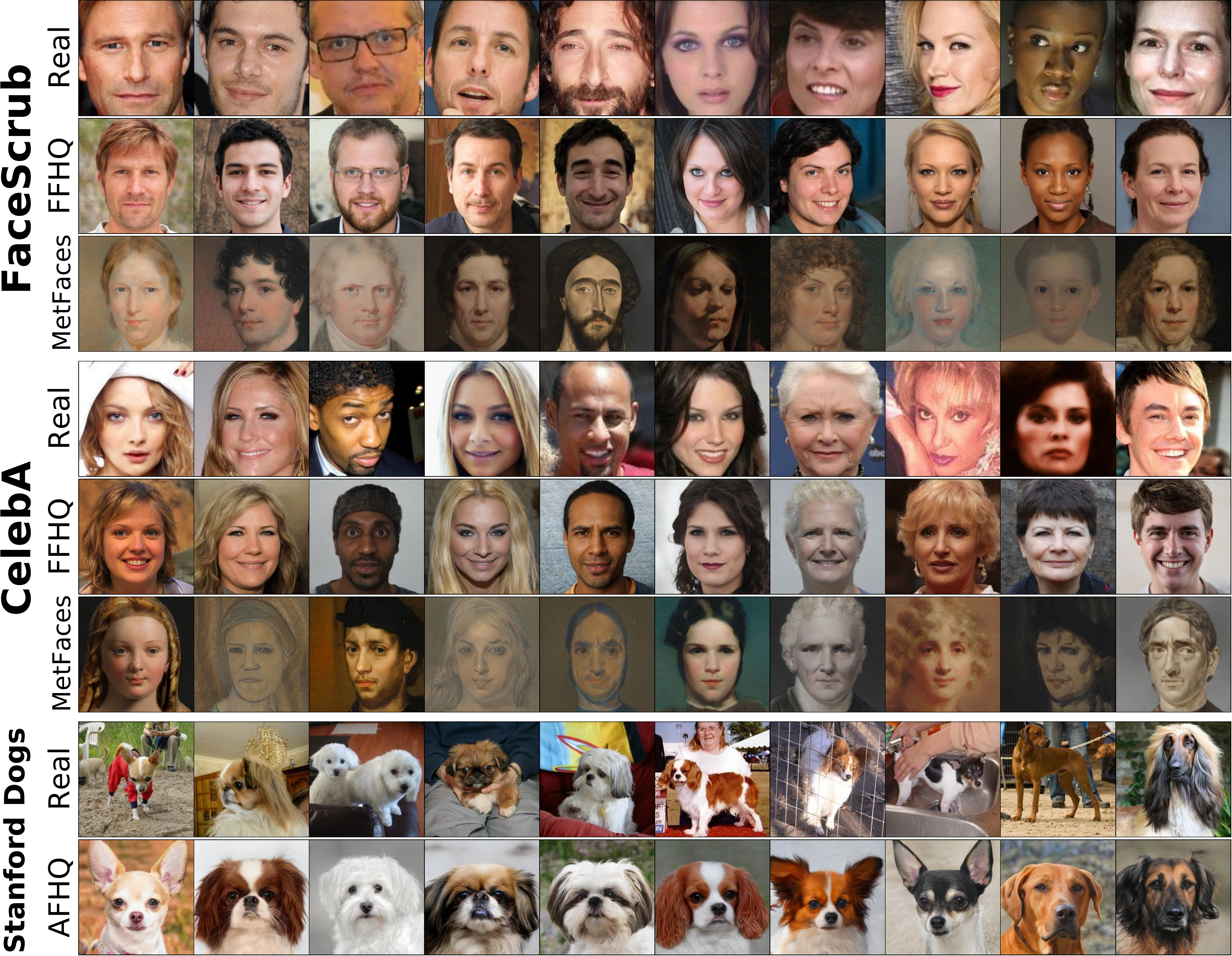}
\caption{Attack results for our attacks performed with various StyleGAN2 models and ResNeSt-101 target models trained on FaceScrub, CelebA, and Stanford Dogs, respectively. The real samples were hand-picked from the training data to represent the targeted identities. The attack samples were selected without cherry-picking, using our transformation-based selection process. We took the samples with the highest robust prediction scores on the target model. The attacks against the FaceScrub model targeted the first five actors and actresses, respectively. For CelebA and Stanford Dogs, we took the first ten classes. The real samples from the Stanford Dogs dataset were cropped to highlight the targeted dog breeds. However, the uncropped samples used to train the target models contained more background and scenery. Note that most samples from the training data have a much lower resolution than our attack results, illustrating that our attack setting is able to produce characteristic images with higher quality than the original data provides.}
\label{fig:attack_samples_large}
\end{figure*}

We also state a large version of \cref{fig:attack_samples} in \cref{fig:attack_samples_large}. For FaceScrub, we visualized the attack results against the first five actors and actresses. For CelebA and Stanford Dogs, we visualized the first ten classes. Note that we did not cherry-pick the samples but used our transformation-based selection process introduced in \cref{sec:sample_selection_process}. In each case, we took the sample with the highest average prediction score on the target model. The evaluation models were not involved in the selection process.

\begin{figure*}[ht]
\centering
\includegraphics[width=\linewidth]{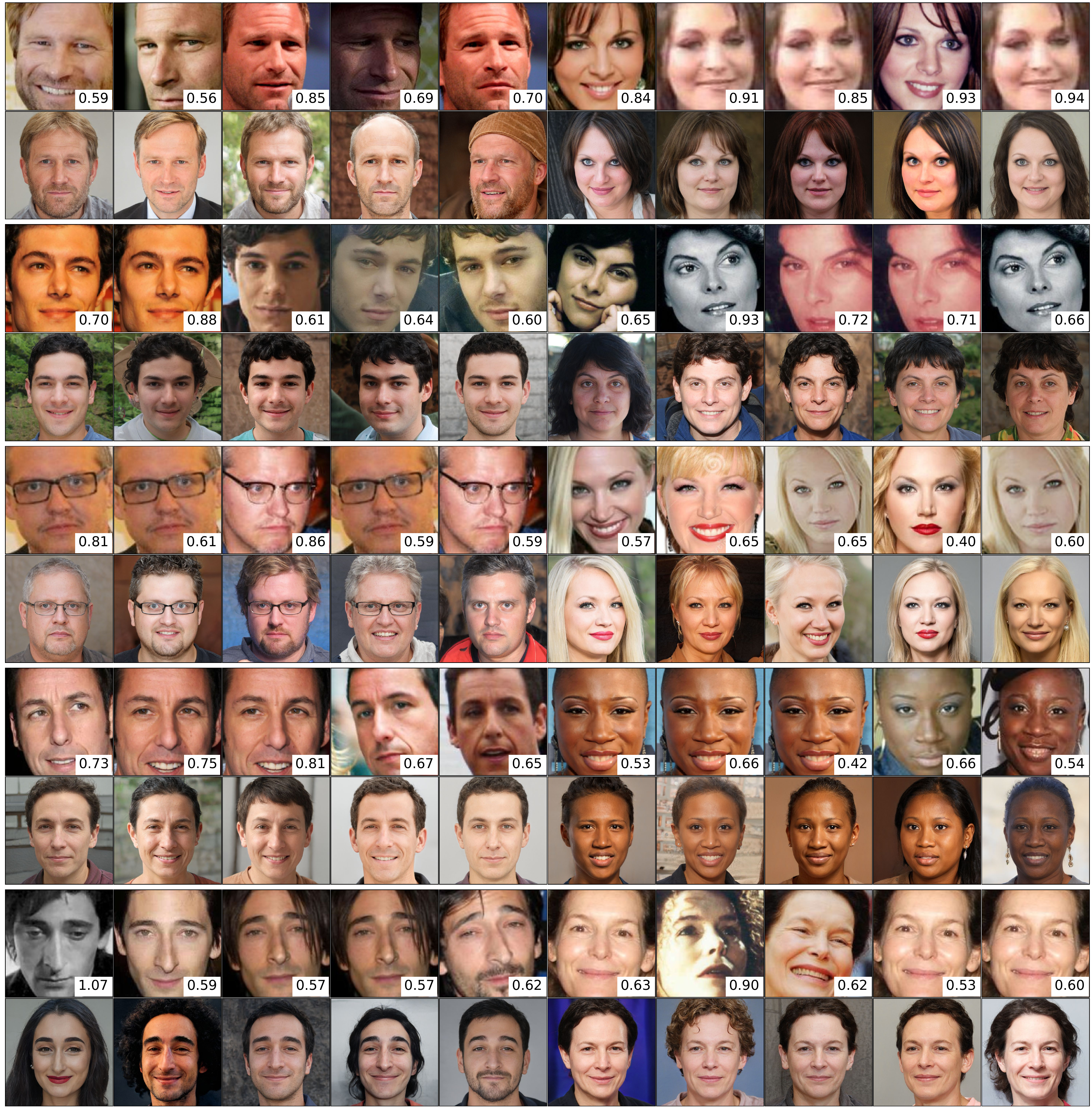}
\caption{Attack results for our attack performed with the pre-trained FFHQ model and the ResNeSt-101 target model trained on FaceScrub. The attacks targeted the first five actors and actresses. All attack samples (bottom rows) were randomly selected. We further plot the nearest neighbors (top rows) from the target model's training data in terms of FaceNet feature distance. We also state the computed feature distance for each image pair.}
\label{fig:add_facescrub_results}
\end{figure*}

We further visualize more samples from our attack against the ResNeSt-101 trained on FaceScrub in \cref{fig:add_facescrub_results} to provide a better overview over the variety and quality of the generated examples. Samples were randomly selected from the attack results, attacking the first five actor and actresses. We plot each image together with its nearest neighbor from the training data by the feature distance from FaceNet. We also state the feature distance for each image pair.

\newpage

\newpage

\newpage

